\documentclass{article}

    \PassOptionsToPackage{numbers, compress}{natbib}
\usepackage[preprint]{neurips_2026}

\usepackage[utf8]{inputenc} % allow utf-8 input
\usepackage[T1]{fontenc}    % use 8-bit T1 fonts
\usepackage{hyperref}       % hyperlinks
\usepackage{url}            % simple URL typesetting
\usepackage{booktabs}       % professional-quality tables
\usepackage{amsfonts}       % blackboard math symbols
\usepackage{nicefrac}       % compact symbols for 1/2, etc.
\usepackage{microtype}      % microtypography
\usepackage{xcolor}         % colors
\usepackage{amsmath}
\usepackage{graphicx}
\usepackage{multirow}
\usepackage{makecell}
\usepackage{cellspace}
\usepackage{tcolorbox}
\tcbuselibrary{breakable,listings}

\newtcblisting{plainpromptbox}{
  breakable,
  listing only,
  colback=black!4,
  colframe=black!25,
  boxrule=0.4pt,
  arc=1pt,
  left=8pt,
  right=8pt,
  top=8pt,
  bottom=8pt,
  listing options={
    basicstyle=\ttfamily\footnotesize,
    breaklines=true,
    columns=fullflexible,
    keepspaces=true,
    showstringspaces=false
  }
}

\makeatletter
\let\@startpbox@action\@startpbox
\makeatother

\title{ConceptTree: Bringing Semantic Transparency to Black-Box Decision Making for Robotic Manipulation}

\author{%
  Yongyan Wen$^1$, Feifan Liu$^1$, Jinyi Chen$^1$, Bo An$^2$, Peng Liu$^1$, Siyaun Li$^{1,\dagger}$\thanks{$^\dagger$Corresponding author.}\\
  $^1$Harbin Institute of Technology, $^2$Nanyang Technological University\\
  \texttt{siyuanli@hit.edu.cn}\\
}

\begin{document}

\maketitle

\begin{abstract}
  Establishing interpretable decision-making processes in long-horizon robotic manipulation is critical for enabling reliable human oversight and intervention. However, existing approaches to robotic manipulation largely treat skill selection as opaque mappings from observations to actions, offering limited transparency into how decisions are formed. In this work, we propose ConceptTree, a framework that reframes high-level manipulation skill selection as reasoning over human-interpretable concepts, representing high-level policies as a sequence of concept-level predicates over visual observations. Rather than relying on implicit latent representations, our method learns a normalized concept space grounded in visual inputs, over which a decision tree is trained to predict high-level skills. This formulation yields a transparent decision process that is both traceable and intervenable, enabling direct inspection and modification of policy behavior. We evaluate our approach on a set of real-world robotic manipulation tasks with increasing complexity. Experimental results show that ConceptTree consistently outperforms existing concept-based baselines, particularly in complex, long-horizon scenarios. Furthermore, we provide qualitative case studies showing that our model supports fine-grained intervention by modifying individual concepts, enabling targeted correction of decision errors without retraining.
\end{abstract}

\section{Introduction}
Recent advances in robotic manipulation have enabled robots to perform increasingly complex tasks in real-world environments \cite{zitkovich2023rt,kim2025openvla,black2025pi}, such as object rearrangement \cite{ding2023task}, assembly \cite{luo2025precise}, and tool-use tasks \cite{lu2024dynamic}. Completing these tasks often involves long-horizon decision making and requires reasoning over continuously changing visual scenes \cite{kroemer2021review}, where success depends on selecting the appropriate high-level manipulation skill based on the current observation. To handle such complexity, a common paradigm is to decompose decision making into multiple levels of abstraction, analogous to hierarchical control in biological systems, where high-level planning and low-level execution are separated. In robotics, this is often instantiated as hierarchical policies \cite{bai2025towards}, where a high-level module selects discrete skills based on observations, and low-level controllers execute them. Without an interpretable decision process, an incorrect skill choice can lead to failed trajectories, where the delayed failures are hard to diagnose.

Despite their strong performance, existing high-level decision approaches primarily rely on end-to-end neural policies or vision-language model-based policies, where decisions are produced by black-box systems \cite{wu2023embodied,song2023llm,brohan2023can,du2024video}. While these models are effective at capturing complex patterns, the reasoning behind their decisions remains implicit. As a result, it is often unclear why a particular skill is selected given an observation, making it difficult to interpret the process and debug the system. This lack of transparency is particularly problematic in real-world deployment, where failures must be diagnosed and corrected in a reliable and expected manner. Without transparency in the decision-making process, when a robot selects an incorrect skill, it is challenging to determine whether the failure arises from perception errors, ambiguous scene understanding, or flawed decision logic, limiting the ability to correct the system systematically. Beyond debugging, transparent decision making also supports human trust in robotic systems by allowing users to inspect and understand the system's behavior \cite{sobrin2025generating}.

While interpretability has been extensively studied in reinforcement learning and sequential decision making \cite{hassija2024interpreting,glanois2024survey,milani2024explainable}, interpretable high-level decision making for visual robotic manipulation remains relatively underexplored. Visual robotic manipulation introduces a distinct challenge: high-level skill choices are governed by physical preconditions and effects, such as whether an object is inside a container, or a door is open. An explanation must therefore reveal not only which visual features influence a decision, but how perceived scene states justify the selected skill in the evolving manipulation process. The closest relevant methods provide useful ingredients but do not fully address this need. Inherently interpretable methods, such as symbolic policies \cite{edmonds2019tale,luo2024end} and decision trees \cite{wen2025skilltree}, make decision structures explicit, but typically rely on low-dimensional states or structured inputs rather than raw visual observations. Post-hoc explanation methods \cite{greydanus2018visualizing,puri2020explain} can be applied to black-box policies, but they often identify influential image regions or features after training without exposing the decision logic that maps perceived scene properties to skill choices. Recent approaches that combine neural perception with structured decision models partially narrow this gap \cite{tanno19adaptive,wan2021nbdt,marton2025mitigating}, yet their rules are often defined over latent features, making them difficult to inspect or intervene on in terms of task-relevant scene properties. These limitations point to a missing interface for robotic manipulation: high-level decisions should be grounded in visual observations through human-interpretable concepts that capture task-relevant object states and spatial relations, while the decision process over these concepts should remain explicit and traceable.

To address these limitations, we propose ConceptTree, a framework that combines semantically grounded concept learning with structured decision making for robotic manipulation. Its concept layer maps visual observations to manipulation-specific concepts, such as object states and spatial relations, forming a semantic interface for high-level skill reasoning. We construct concept-level annotations with vision-language models (VLMs) as verifiers of task-relevant object-state and spatial-relation predicates, enabling concept-layer training without manual concept labels. At inference time, the learned concept values are passed to a decision tree, which selects skills through inspectable decision paths. The resulting framework makes high-level decisions through a sequence of predicates over task-relevant concepts, so each prediction can be traced as an explicit decision path rather than an opaque output. We evaluate ConceptTree on four real long-horizon manipulation tasks with varying temporal dependencies and complexity. The results show that ConceptTree achieves the strongest overall completion rates among the compared methods, with particularly clear gains on more complex tasks where simple concept classifiers and black-box VLM policies degrade. Ablation studies further show that reliable concept supervision and temporal context improve decision quality while yielding compact trees with short concept-level decision paths. Finally, qualitative intervention results demonstrate that decision errors can be localized to specific concept values and corrected by modifying individual concepts without retraining.
% Our contributions are summarized as follows:
% \begin{itemize}
%   \item We introduce VLM-verified concept learning for visual robotic manipulation, mapping raw observations to task-relevant predicates such as object states and spatial relations.
%   \item We propose ConceptTree, a tree-structured high-level policy that selects skills over learned concept activations and produces explicit visual-to-skill decision paths.
%   \item We validate ConceptTree on real long-horizon manipulation tasks, showing stronger skill prediction, analyzing the effects of concept supervision and tree structure, and demonstrating concept-level error localization and intervention.
% \end{itemize}

\section{Related Work}
\textbf{Interpretability for Robotic Manipulation} Interpretability in robotic manipulation is closely related to broader efforts in interpretable machine learning \cite{hassija2024interpreting} and sequential decision making \cite{glanois2024survey,milani2024explainable}. One line of work focuses on learning inherently interpretable policies, such as policy distillation \cite{bastani2018verifiable, kohler2024interpretable}, symbolic policies \cite{edmonds2019tale,luo2024end}, linear models \cite{wabartha2024piecewise}, or decision trees \cite{tanno19adaptive,dhebar2020interpretable,paleja2022learning,dhebar2022toward,wen2025skilltree}, where the decision process is explicitly represented. These approaches provide transparent reasoning and allow direct inspection of decision rules. However, they typically rely on low-dimensional state representations or structured inputs. Another line of work adopts post-hoc explanation methods to interpret black-box policies \cite{greydanus2018visualizing,puri2020explain}. These approaches aim to explain decisions after training, often by identifying input regions or features that influence action selection. While flexible and widely applicable, such methods may not faithfully reflect the internal decision process of the model and are limited in their ability to support reasoning, verification, or intervention. In complex manipulation tasks involving multiple objects and temporal dependencies, this gap becomes more pronounced, as explanations may not align with the underlying decision logic. Recent efforts that integrate neural perception with structured decision models partially address this gap \cite{wan2021nbdt,marton2025mitigating}, but typically construct decision rules over latent features rather than semantically meaningful variables. In contrast, our approach performs decision making directly over concept representations that are explicitly grounded in the visual scene. By combining reliable concept supervision with a structured decision mechanism, our method enables interpretable, verifiable, and intervenable reasoning in high-dimensional, sequential manipulation tasks.

\textbf{Interpretable Representation} Concept-based interpretability introduces human-interpretable intermediate variables (e.g., concepts) to expose the internal reasoning of neural networks \cite{lee2024concept}. A representative paradigm is the concept bottleneck model (CBM) \cite{koh2020concept}, which decomposes prediction into concept inference followed by decision making over the concept space. This formulation provides a structured interface for inspection and intervention, and has been widely adopted as a foundation for interpretable learning. However, despite these advantages, existing CBM-based approaches share several fundamental limitations. First, concept representations are often either manually annotated or approximated through pretrained models, which can lead to a mismatch between the defined concepts and the actual visual properties in the environment. In particular, similarity-based supervision with pretrained models (e.g., CLIP \cite{radford2021learning} or GroundingDINO \cite{liu2024grounding}) commonly used in label-free settings \cite{oikarinen2023label} does not guarantee that a concept is truly present in the scene, making it difficult to reliably capture task-relevant properties such as object states or spatial relations. Moreover, CBMs typically employ simple decision functions over concept representations, most commonly linear classifiers \cite{yuksekgonul2023post,srivastava2024vlg,hu2025semi,liu2025hybrid}. In such formulations, interpretability is primarily reflected through concept weights, which provide limited insight into how multiple concepts interact to produce a decision. As a result, while individual concepts are interpretable, the overall decision process remains implicit and lacks a clear reasoning structure. Moreover, most CBM-based methods are developed and evaluated in classification settings, where concepts are relatively stable and well-aligned with semantic categories. In contrast, embodied manipulation tasks introduce additional challenges, including dynamic scenes, fine-grained object interactions, and strong temporal dependencies. These factors require concept representations that are not only semantically meaningful but also sufficiently reliable to support downstream decision making in structured environments.

\section{ConceptTree}
\begin{figure}[htbp]
  \centering
  \includegraphics[width=\textwidth]{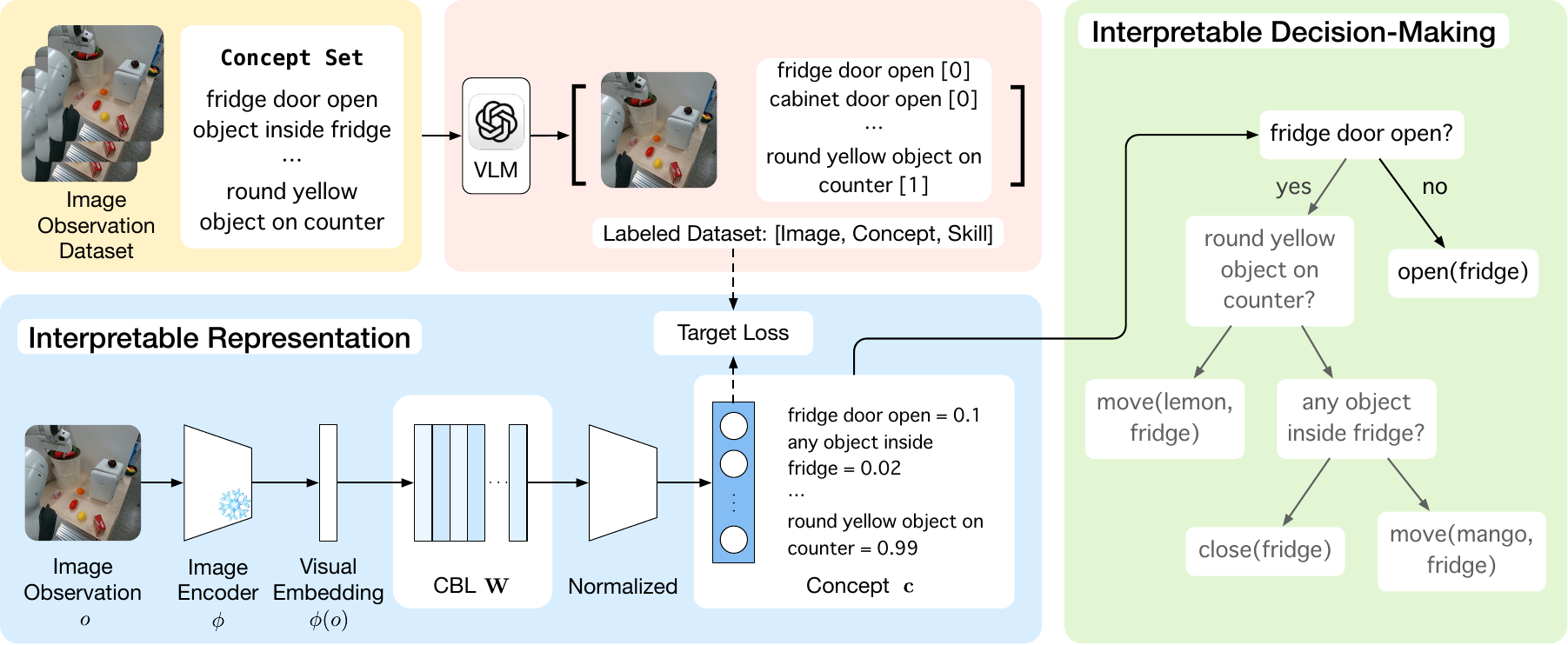}
  \caption{\textbf{Overview of the ConceptTree framework.} The model consists of three main components: (i) a visual encoder that extracts features from observations, (ii) a concept projection layer that maps features to a structured concept representation, and (iii) a decision tree that predicts high-level skills based on the concept vector. During training, a VLM provides binary concept judgments for supervision, while at inference time, the system operates without access to the VLM, relying solely on the learned concept representation and decision tree.}
  \label{fig.framework}
\end{figure}

In this section, we present ConceptTree, a novel framework for interpretable high-level policy learning that unifies semantically grounded representation learning with structured decision making. Our approach decomposes the policy into two complementary components: a concept layer that maps visual observations to interpretable representations (concepts), and a decision module that performs reasoning over these concepts to select skills. This design allows decisions to be interpreted at both the representation and decision levels. Specifically, concept values provide a human-interpretable semantic description of the perceived scene, while the decision tree exposes the reasoning process that leads to the final skill. The overall pipeline is illustrated in Fig. \ref{fig.framework}.

% \subsection{Problem Formulation}\label{ssec.problem}
We consider high-level decision making in long-horizon robotic manipulation under a hierarchical control framework. At each time step $t$, the agent selects a discrete skill $a_t\in \mathcal{A}$ from a predefined task-relevant skill set $\mathcal{A}$, which is then executed by a low-level policy. Our focus is on learning an interpretable high-level policy that maps visual observations to skill decisions. At time step $t$, the agent receives an RGB observation $o_t\in \mathbb{R}^{H\times W\times 3}$. Instead of directly mapping observations to skills, we introduce an intermediate concept representation that captures semantically meaningful properties of the scene. Specifically, each observation is mapped to a concept vector $\mathbf{c}_t\in \mathbb{R}^M$, where each dimension corresponds to a task-relevant concept (e.g., object states or spatial relations). To model short-term temporal dependencies, we optionally incorporate context from the previous step. Rather than feeding multiple frames into the perception module, we maintain a modular design in which each observation is processed independently into its concept representation. Temporal information is then incorporated at the decision level by combining concept vectors from consecutive steps. The effective input to the policy is therefore defined as
\begin{equation}
  \tilde{\mathbf{c}}_t =
    \begin{cases}
      \mathbf{c}_t, & \text{w/o history} \\
      [\mathbf{c}_{t-1}, \mathbf{c}_t], & \text{w/ history}
    \end{cases}
\end{equation}

The high-level policy $\pi_\theta(a_t|\tilde{\mathbf{c}}_t)$ is thus formulated as a mapping from concept representations to actions. This formulation separates perception and decision making into two interpretable stages. The concept representation $\mathbf{c}_t$ provides a semantically grounded description of the scene, while the policy $\pi_\theta$ defines an explicit reasoning process over these concepts. This decomposition enables decisions to be analyzed, verified, and modified at the level of individual concepts, facilitating interpretable and controllable policy behavior.

\subsection{Concept Supervision via VLM Annotation}\label{ssec.concept_supervision}
To learn semantically grounded concept representations without manual annotation, we construct concept-level supervision using a VLM (e.g., GPT-5.1) as an external annotator. We first obtain a predefined task-relevant concept set $\mathcal{C} = \{c^{(1)},\dots,c^{(M)}\}$ from a large language model (LLM), which generates task-relevant concepts from high-level task specifications (see Appendix \ref{sec.concept_construction} for more details). Here, $c^{(k)}$ denotes the $k$-th semantic concept in the predefined concept set. Given a dataset of observations and ground-truth skill labels $\mathcal{D} = \{(o_t, a_t^*)\}_{t=1}^T$, where $a_t^*$ denotes the expert skill at time step $t$, we associate each observation $o_t$ with a predefined set of $M$ task-relevant concepts. For each concept $i\in \{1,\dots,M\}$, we define a binary predicate over the observation, indicating whether the concept holds in the scene. We use a VLM to approximate these predicates. Specifically, for each observation $o_t$ and concept $\mathbf{c}\in \mathcal{C}$, the VLM produces a binary judgment:
\begin{equation}
  y_t(c^{(k)})\in \{0,1\},
\end{equation}
where $y_t(c^{(k)}) = 1$ indicates that concept $c^{(k)}$ is present in $o_t$, and $0$ otherwise. Collectively, this yields a concept label vector $\mathbf{y}_t=\big(y_t(c^{(1)}),\dots,y_t(c^{(M)})\big)^\top\in \{0,1\}^M$ for each observation. This process defines an augmented dataset $\mathcal{D}_c = \{(o_t,\mathbf{y}_t,a_t^*)\}_{t=1}^T$, which provides supervision for both concept learning and downstream decision making.

Importantly, the VLM is used only during dataset construction and is not required at inference time. This can be viewed as a form of supervision distillation, where concept-level reasoning from a large model is transferred into a lightweight and structured representation. Compared to similarity-based supervision, which produces continuous scores, the proposed verification scheme provides explicit binary signals aligned with task semantics, enabling more reliable supervision for concepts involving object states and spatial relations.

\subsection{Concept Learning via Projection and Task-Aware Supervision}
Given the concept supervision defined in Section \ref{ssec.concept_supervision}, we learn a concept representation that maps visual observations to semantically grounded concept values.

\paragraph{Concept Projection.}
Each observation $o_t$ is first encoded into a visual feature vector: $h_t = \phi(o_t), h_t \in \mathbb{R}^d$, where $\phi$ denotes a fixed visual encoder. We then project the feature into the concept space $z_t = \mathbf{W} h_t, z_t \in \mathbb{R}^M$, where $\mathbf{W} \in \mathbb{R}^{M\times d}$ maps visual features to $M$ concept dimensions. To obtain normalized concept values, we apply a per-dimension calibration:
\begin{equation}
  \mathbf{c}_t^{(k)} = \sigma(p_k z_t^{(k)}+q_k),
\end{equation}
where $\sigma(\cdot)$ is the sigmoid function, $(p_k, q_k)$ are learnable parameters for each concept, and $\mathbf{c}_t^{(k)}\in (0, 1)$ denotes the learned concept representation predicted from observation $o_t$. Using the binary concept labels $y_t(c^{(k)})$ obtained from VLM verification, we train the concept representation with a binary cross-entropy loss:
\begin{equation}
  \mathcal{L}_{\text{concept}} = \sum_{t=1}^T \text{BCE}(\mathbf{c}_t, \mathbf{y}_t),
\end{equation}
where $\mathbf{c}_t=\big(\mathbf{c}_t^{(1)},\dots,\mathbf{c}_t^{(M)}\big)^\top$ is the learned concept representation.

\paragraph{Skill-Aware Regularization.}
In addition to concept supervision, we introduce a skill-aware objective to encourage the learned representation to retain information relevant for decision making. Specifically, we attach a linear classifier $f(\mathbf{c}_t)$ on top of the concept vector and optimize:
\begin{equation}
  \mathcal{L}_{\text{skill}} = \sum_{t=1}^T \text{CE}(f(\mathbf{c}_t), a_t^*),
\end{equation}
where $a_t^*$ is the ground-truth skill label. The overall objective for concept learning is:
\begin{equation}
  \mathcal{L}_{\text{target}} = \mathcal{L}_{\text{concept}} + \lambda \mathcal{L}_{\text{skill}},
\end{equation}
where $\lambda$ controls the trade-off between semantic alignment and skill relevance. The resulting concept layer is not used as a complete predictor by itself. Rather, it serves as an intermediate semantic interface for the high-level policy, which is instantiated with a decision tree to obtain a transparent decision process.

\subsection{Concept-based Policy Learning with Decision Tree}
Given the learned concept representation $\mathbf{c}_t$, we perform high-level decision making by explicitly reasoning over the concept space. We model the high-level policy as a function $g: \mathbb{R}^{\tilde M} \rightarrow \mathcal{A}$ that maps concept representations to skill decisions $a_t = g(\tilde{\mathbf{c}}_t)$, where $\tilde M = M$ or $2M$ depending on whether temporal context (history) is used.

Specifically, we instantiate $g$ as a decision tree, where each internal node corresponds to a binary predicate over a single concept dimension:
\begin{equation}
  d_n(\tilde{\mathbf{c}}_t)=\mathbb{I}\!\left[\tilde{c}_t^{(k)}\le \tau_n\right],
\end{equation}
where $k$ indexes a concept dimension and $\tau_n$ is a learned threshold. If $d_n(\tilde{\mathbf{c}}_t)=1$, the sample follows the negative (left) branch of node $n$; otherwise, it follows the positive (right) branch. A decision corresponds to traversing a path from the root to a leaf node, resulting in a sequence of concept-level predicates:
\begin{equation}
  \bigwedge_{n \in \mathcal{P}} d_n(\tilde{\mathbf{c}}_t),
\end{equation}
where $\mathcal{P}$ denotes the set of nodes along the decision path. This formulation yields an explicit and interpretable reasoning process, where each decision is explained by a conjunction of concept-based conditions. The decision tree is then trained from the concept representation, allowing the reasoning structure to operate over fixed, semantically grounded inputs. Unlike a linear classifier \cite{wong2021leveraging} that explains predictions through global weights, the tree assigns each skill prediction to a local sequence of semantic predicates. This makes the selected skill directly traceable to the specific scene conditions tested along the path.

\paragraph{Interpretability and Intervention.}
In the formulation, each prediction can be traced to a sequence of concept-level predicates, providing a transparent explanation of the decision. Furthermore, the explicit structure supports targeted intervention. Given a modified concept vector $\tilde{\mathbf{c}}_t'$, the decision outcome can be directly recomputed as $a_t' = g(\tilde{\mathbf{c}}_t')$, allowing one to analyze how changes in specific concepts affect the final decision. This enables fine-grained diagnosis and correction of errors without modifying the underlying model parameters. In contrast to CBMs, which typically use simple linear classifiers over concept representations, our approach performs structured reasoning through a tree-based decision process. This explicitly separates representation learning from decision making, enabling both semantically grounded perception and transparent, verifiable reasoning.

\section{Experiment Results}
In this section, we evaluate our approach on real long-horizon robotic manipulation tasks to assess both decision performance and interpretability, focusing on how semantically grounded representations and structured decision processes contribute to reliable and controllable behavior.

\begin{figure}
  \centering
  \includegraphics[width=\textwidth]{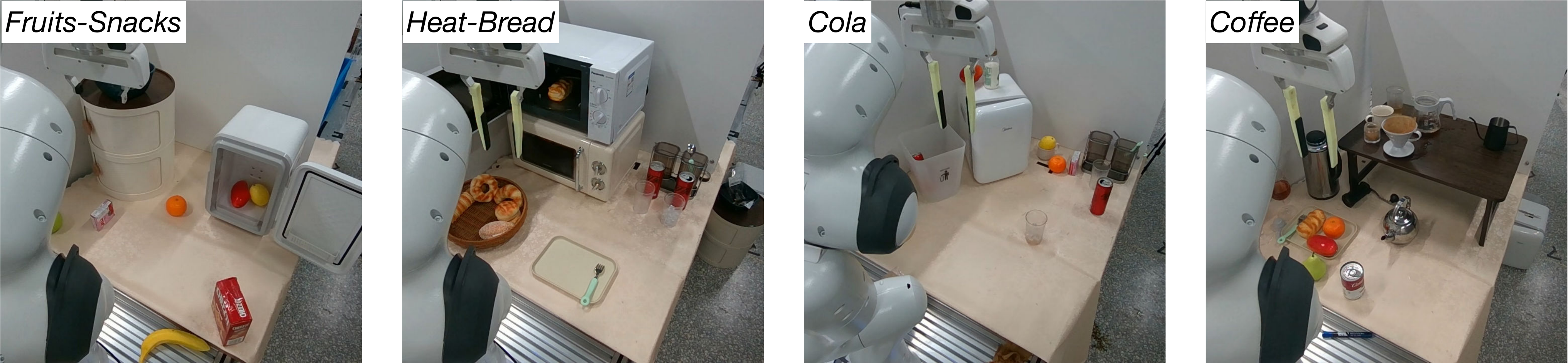}
  \caption{\textbf{Tasks evaluated in our study}, including storing fruits and snacks, heating and preparing the bread, preparing a cup of ice cola and storing the milk, manual coffee brewing, with example observations.}
  \label{fig.tasks}
\end{figure}

\subsection{Experimental Setup}
\paragraph{Tasks.}
We evaluate our approach on four long-horizon robotic manipulation tasks with progressively increasing complexity: \textit{Fruits-Snacks}, \textit{Heat-Bread}, \textit{Cola}, and \textit{Coffee}, as illustrated in Fig. \ref{fig.tasks}. These tasks are designed to systematically vary along three key dimensions: temporal dependency, object interaction complexity, and sensitivity to visual details. Specifically,
\begin{itemize}
  \item \textit{Fruits-Snacks} serves as a baseline task with a fixed sequence of object placements and minimal temporal dependency, where decisions are largely determined by the current observation.
  \item \textit{Heat-Bread} introduces temporal dependency through skill reuse, requiring the policy to reason about previously executed actions.
  \item \textit{Cola} further increases difficulty by involving smaller objects and finer spatial interactions, making accurate perception more critical.
  \item \textit{Coffee} represents the most challenging setting, combining temporal dependency with complex object interactions and sensitivity to subtle visual cues.
\end{itemize}

\paragraph{Baselines.}
We compare our approach with four representative baselines: (i) an end-to-end \textbf{VLM policy} that directly maps observations to skills according to the task description as a black-box model; (ii) \textbf{Label-free CBM} \cite{oikarinen2023label}, a CLIP-based concept model trained with similarity-based supervision; (iii) \textbf{VLG-CBM} \cite{srivastava2024vlg}, a VLM-supervised concept model with a linear decision head. The concept-based baselines share the same concept sets as our method. We additionally evaluate (iv) \textbf{ConceptSLC}, which uses the same learned concept representation as ConceptTree but replaces the decision tree with the sparse linear classifier \cite{wong2021leveraging} used in \cite{oikarinen2023label}. This variant isolates the effect of the decision module and tests whether tree-structured reasoning provides benefits beyond sparse linear prediction over the same concepts.

\paragraph{Evaluation Metrics.}
In long-horizon tasks, intuitively, an early wrong skill usually indicates a fundamental decision failure, whereas a late error may occur after most high-level decisions have already been made correctly. We evaluate high-level policies on real robot manipulation episodes using completion rate (CR), which measures how far a predicted skill sequence progresses before the first skill decision error. For an episode with length $T$, CR is defined as
\begin{equation}
  \text{CR} = \frac{1}{T}\max_{m} \{a_t = a_t^*, \forall t \le m\}.
\end{equation}

\begin{table}[t]
\centering
\caption{\textbf{Completion rate comparison on four tasks}. We run 5 seeds for each method (mean $\pm$ std in \% and H means history). The maximum depth is set to 7 for ConceptTree across all tasks.}
\label{tbl.main_results}
  \setlength{\tabcolsep}{4pt}
  \begin{tabular}{c Sl Sc Sc Sc Sc}
    \toprule
    \multirow{2}{*}{Category} & \multirow{2}{*}{Method} & \multicolumn{4}{c}{Task} \\
    \cmidrule{3-6}
    & & \textit{Fruits-Snacks} & \textit{Heat-Bread} & \textit{Cola} & \textit{Coffee} \\
    \cmidrule{1-6}
    \multirow{5}{*}{w/o H}
    & VLM Policy & $42.50\pm1.24$ & $40.00\pm1.75$ & $21.33\pm0.93$ & $5.09\pm2.37$ \\
    & Label-free CBM & $7.22\pm2.06$ & $6.29\pm1.28$ & $15.11\pm2.79$ & $9.82\pm3.77$ \\
    & VLG-CBM & $63.89\pm2.60$ & $15.43\pm6.34$ & $13.56\pm1.45$ & $13.82\pm4.91$ \\
    & ConceptSLC & $85.71\pm0.00$ & $29.71\pm1.40$ & $69.78\pm2.47$ & $14.55\pm6.50$ \\
    \cmidrule{2-6}
    & ConceptTree (Ours) & $\mathbf{95.56}\pm2.54$ & $\mathbf{35.71}\pm7.00$ & $\mathbf{78.44}\pm5.60$ & $\mathbf{22.55}\pm14.48$ \\
    \cmidrule{1-6}
    \multirow{5}{*}{w/ H}
    & VLM Policy & $43.06\pm2.78$ & $68.00\pm2.39$ & $19.56\pm0.99$ & $5.45\pm2.23$ \\
    & Label-free CBM & $34.44\pm9.29$ & $11.71\pm3.56$ & $35.78\pm4.11$ & $13.45\pm3.04$ \\
    & VLG-CBM & $56.39\pm17.22$ & $28.86\pm3.26$ & $24.67\pm5.63$ & $17.09\pm4.19$ \\
    & ConceptSLC & $\mathbf{100.00}\pm0.00$ & $79.71\pm3.88$ & $87.33\pm2.18$ & $18.18\pm6.08$ \\
    \cmidrule(lr){2-6}
    & ConceptTree (Ours) & $94.92\pm6.22$ & $\mathbf{84.57}\pm4.64$ & $\mathbf{95.11}\pm4.48$ & $\mathbf{37.09}\pm17.36$ \\
    \bottomrule
  \end{tabular}
\end{table}

\subsection{Main Results}
Unless otherwise specified, we use a fixed maximum tree depth of 7 for ConceptTree, which provides a good balance between decision performance and interpretability according to the ablation study. As shown in Table \ref{tbl.main_results}, ConceptTree achieves the strongest overall completion rates across the evaluated tasks, with particularly clear gains on the more challenging \textit{Heat-Bread}, \textit{Cola}, and \textit{Coffee} tasks. Compared with the black-box VLM policy, ConceptTree is substantially more reliable, suggesting that directly querying a single VLM for high-level skill selection is insufficient for long-horizon manipulation. Among concept-based methods, Label-free CBM and VLG-CBM perform poorly as task complexity increases, indicating that concept bottlenecks alone do not guarantee robust high-level decision making. More importantly, ConceptTree consistently outperforms ConceptSLC on the temporally dependent and visually demanding tasks, although ConceptSLC slightly outperforms it on the easier \textit{Fruits-Snacks} task with history. Since ConceptSLC uses the same learned concept representation as ConceptTree and differs only in the decision module, this comparison shows that the performance gain does not come merely from better concept learning. Instead, structured reasoning over concepts is critical for robust high-level skill selection. As further analyzed in Appendix \ref{sec.dt_vs_linear}, ConceptTree also provides more compact explanations than the sparse linear model, using fewer concepts and shorter decision paths for test-time predictions. Overall, these results support our claim that ConceptTree benefits from making decisions explicitly in the concept space through an interpretable tree-structured process.

\subsection{Ablation Studies}
\begin{figure}[tbp]
  \centering
  \includegraphics[width=\textwidth]{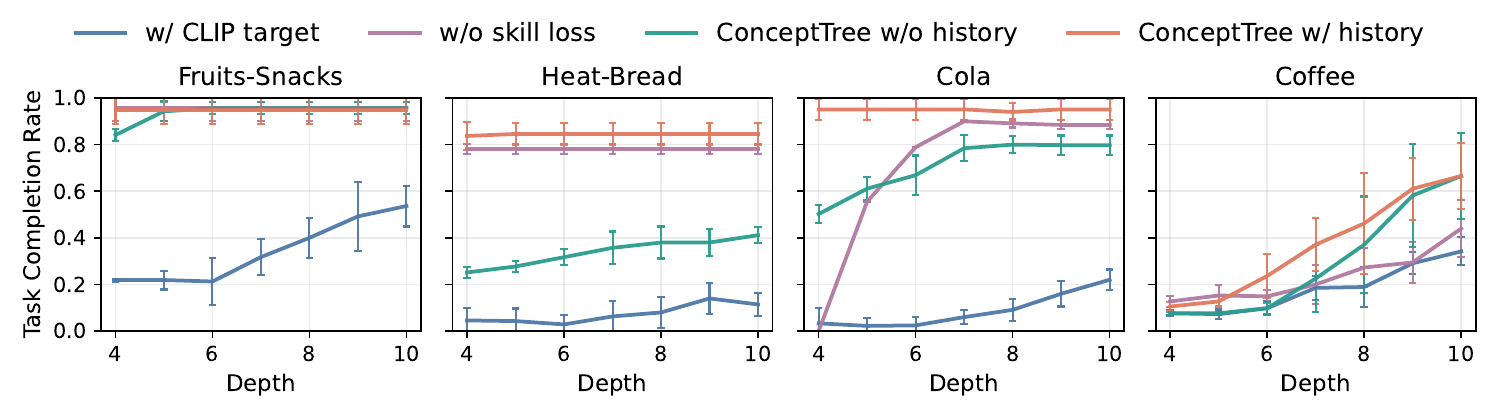}
  \caption{\textbf{Ablation study across four tasks.} Bars indicate standard deviation. Depth denotes the maximum allowed tree depth during training, not the actual depth.}
  \label{fig.ablation}
\end{figure}

\begin{figure}[t]
  \centering
  \includegraphics[width=\textwidth]{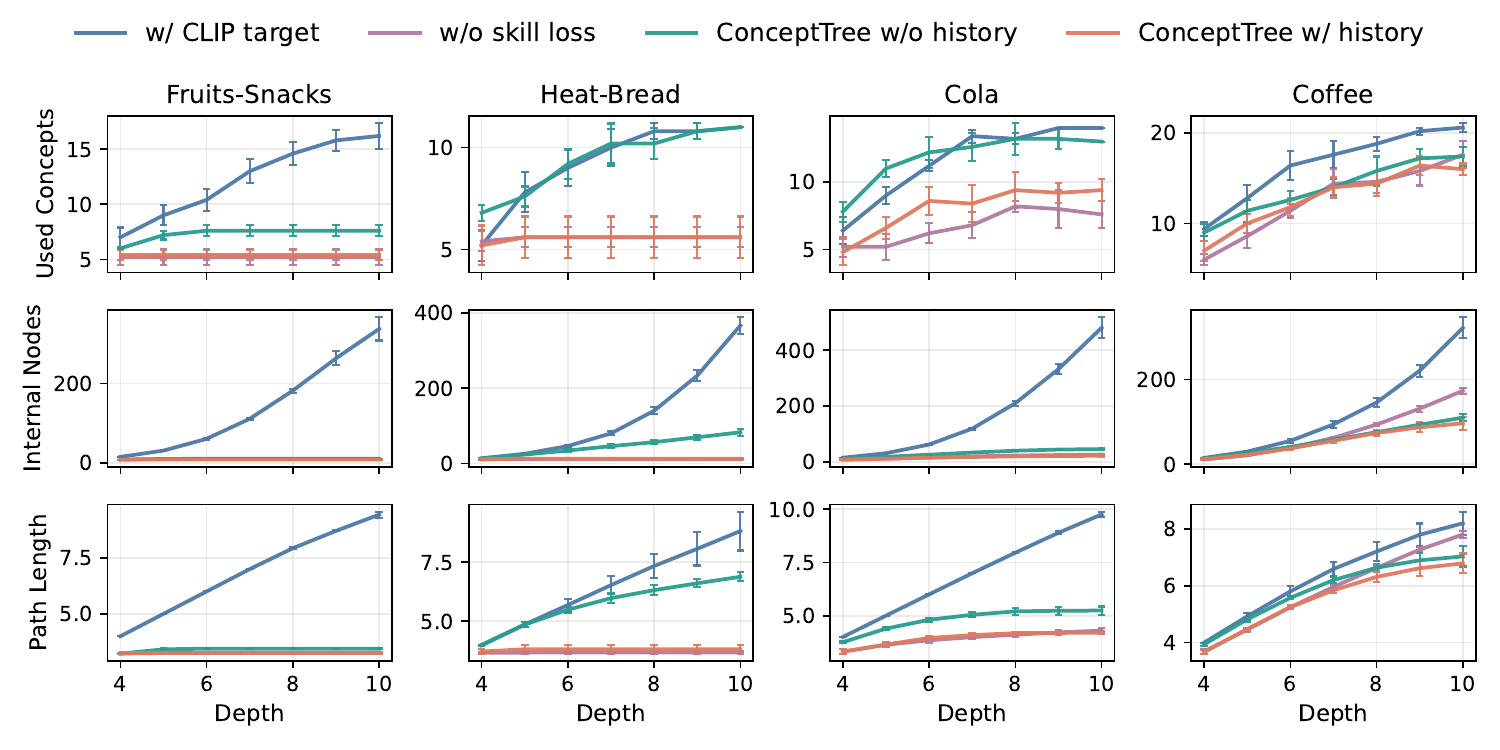}
  \caption{\textbf{Tree complexity under different ablation settings.} We report the number of used concepts and leaf nodes and bars indicate standard deviation.}
  \label{fig.tree_complexity}
\end{figure}

We analyze how different design components affect both decision performance and the structure of the learned decision tree. All ablation studies are conducted offline on a held-out validation set constructed from recorded robot observations. Results are summarized in Fig. \ref{fig.ablation} and Fig. \ref{fig.tree_complexity}. Across all settings, we observe that better performance consistently correlates with simpler decision structures, characterized by shallower trees, fewer concepts, and shorter decision paths.

\textbf{Effect of supervision quality (w/ CLIP target and w/o skill loss).} Concept supervision plays a dominant role in overall performance. Replacing VLM-based supervision with CLIP similarity leads to a substantial drop in performance across all tasks, accompanied by significantly larger trees with increased concept usage and longer decision paths. In contrast, removing the skill loss results in a more moderate performance degradation. These results indicate that the quality of concept supervision is critical for learning compact and effective decision structures, while task-aligned training provides additional but comparatively limited improvements.

\textbf{Effect of temporal context (w/o history).} Removing history input has little effect on simpler tasks such as \textit{Fruits-Snacks}, but leads to clear performance drops in tasks with stronger temporal dependencies like \textit{Heat-Bread} or \textit{Coffee}. In contrast, the tree structure remains largely unchanged, suggesting that temporal information primarily improves decision accuracy under partial observability, rather than altering the structure of the learned policy. The impact of temporal context depends on task properties.

\textbf{Effect of tree depth.} Increasing the maximum tree depth improves performance up to a saturation point, reflecting increased expressive capacity. Notably, ConceptTree achieves strong performance even with very shallow trees (e.g., depth = 4) across most tasks. At this depth, the resulting decision paths are also short (with an average length below 4), indicating that accurate decisions can be made with only a few concept-level predicates. This suggests that the learned policies remain both effective and interpretable, without requiring deep or complex decision structures.

\textbf{Summary.} Across all settings, we observe a consistent relationship between concept quality, structural efficiency, and performance. Stronger concept supervision and task-aware training lead to more compact, selective, and shorter decision paths, whereas weaker supervision results in larger trees with longer and less efficient decision chains. These findings suggest that performance gains arise not only from improved concept representations, but critically from how these concepts are structured into coherent and efficient decision processes.

\subsection{Case Study}
\begin{figure}[t]
  \centering
  \includegraphics[width=\textwidth]{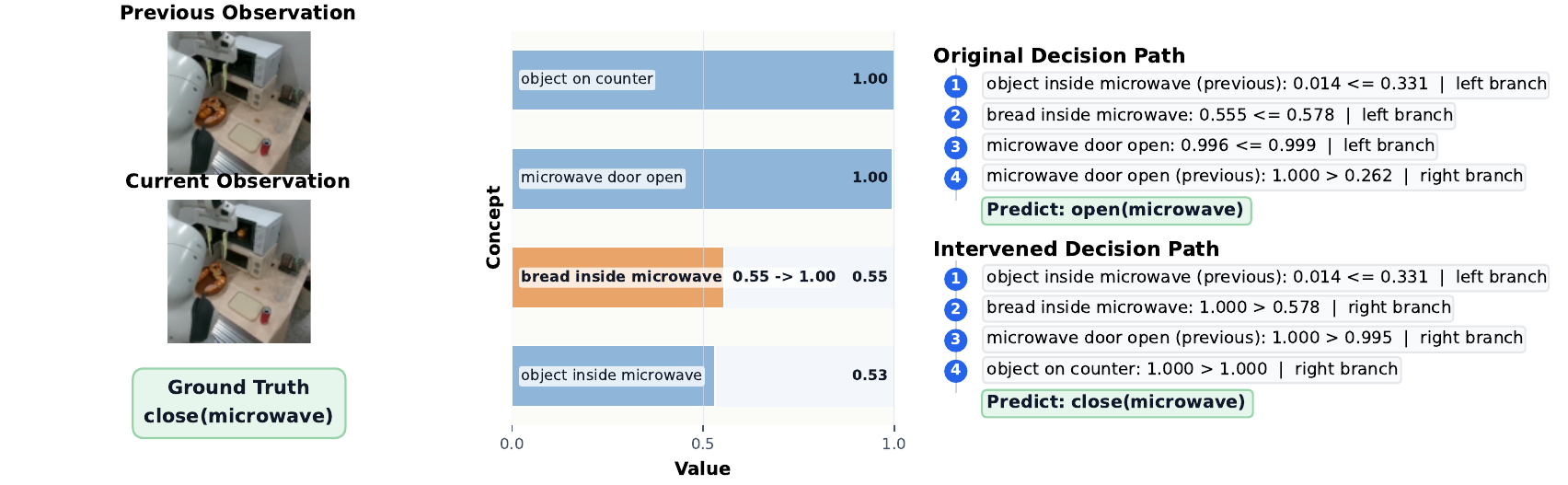}
  \caption{\textbf{Decision-path diagnosis and concept intervention.} In this \textit{Heat-Bread} example, the original path predicts \texttt{open(microwave)} because the value of \textit{bread inside microwave} falls below the split threshold. This is inconsistent with the observation, where the bread is already inside the microwave, so the concept is falsely treated as negative. Manually correcting this single concept value redirects the tree traversal and changes the prediction to the ground-truth skill \texttt{close(microwave)}.}
  \label{fig.intervention}
\end{figure}

To further analyze our method, we present a qualitative case study illustrating the decision process and its response to concept-level intervention. As shown in Fig. \ref{fig.intervention}, we show an example from the \textit{Heat-Bread} task. By inspecting the concept values, we observe that the value of the concept \textit{bread inside microwave} is underestimated (0.555$<$0.578), which leads to an incorrect decision.

The decision tree provides an explicit reasoning path for this prediction. As shown in the original decision path, the low activation value of the concept \textit{bread inside microwave} causes the traversal to follow branches corresponding to the bread not being inside the microwave. This results in selecting the incorrect skill. We then perform an intervention by modifying the concept \textit{bread inside microwave} to 1.0, reflecting the correct semantic state (see identification details in Appendix \ref{sec.additional_intervention}). After the modification, the decision path changes accordingly, and the model predicts the correct skill \texttt{close(microwave)}. Notably, this intervention alters only a single concept value while keeping all other inputs unchanged. This example highlights two key properties of our approach. First, decisions are explicitly grounded in interpretable concept-level predicates, allowing errors to be traced to specific semantic variables. Second, the structured decision process enables fine-grained intervention, where modifying a concept can directly correct the model's behavior without retraining.

\section{Conclusion}
We presented ConceptTree, a framework for interpretable high-level decision making in robotic manipulation that unifies semantically grounded representation learning with structured decision processes. By introducing a concept layer trained with verification-based supervision and a decision tree that performs reasoning over concept representations, our approach enables decisions to be interpreted at both the representation and decision levels. Through experiments on long-horizon manipulation tasks, we showed that ConceptTree achieves strong decision performance while providing transparent and controllable behavior. In particular, our results demonstrate that decision errors can be traced to specific concept dimensions and corrected through targeted intervention, highlighting the practical value of concept-based reasoning for debugging and improving policy behavior. Overall, this work suggests that combining semantically grounded representations with explicit decision structures is a promising direction for building interpretable and reliable robotic systems. Despite these advantages, our approach has several limitations. The quality of concept representations depends on VLM-based supervision, which may be unreliable for subtle visual states or complex spatial relations. Moreover, the method relies on predefined concept and skill sets, which may limit scalability to more open-ended tasks.

\begin{ack}
% Use unnumbered first level headings for the acknowledgments. All acknowledgments

% Do {\bf not} include this section in the anonymized submission, only in the final paper. You can use the \texttt{ack} environment provided in the style file to automatically hide this section in the anonymized submission.
\end{ack}

% \section*{References}

% References follow the acknowledgments in the camera-ready paper. Use unnumbered first-level heading for
% the references. Any choice of citation style is acceptable as long as you are
% consistent. It is permissible to reduce the font size to \verb+small+ (9 point)
% when listing the references.
% Note that the Reference section does not count towards the page limit.
\medskip

\bibliographystyle{plainnat}
\bibliography{ref}

@article{luo2025precise,
    title     = {Precise and dexterous robotic manipulation via human-in-the-loop reinforcement learning},
    author    = {Luo, Jianlan and Xu, Charles and Wu, Jeffrey and Levine, Sergey},
    journal   = {Science Robotics},
    volume    = {10},
    number    = {105},
    pages     = {eads5033},
    year      = {2025},
    publisher = {American Association for the Advancement of Science}
}

@inproceedings{zitkovich2023rt,
    title        = {RT-2: Vision-language-action models transfer web knowledge to robotic control},
    author       = {Zitkovich, Brianna and Yu, Tianhe and Xu, Sichun and Xu, Peng and Xiao, Ted and Xia, Fei and Wu, Jialin and Wohlhart, Paul and Welker, Stefan and Wahid, Ayzaan and others},
    booktitle    = {Conference on Robot Learning},
    pages        = {2165--2183},
    year         = {2023},
    organization = {PMLR}
}

@inproceedings{kim2025openvla,
    title        = {OpenVLA: An Open-Source Vision-Language-Action Model},
    author       = {Kim, Moo Jin and Pertsch, Karl and Karamcheti, Siddharth and Xiao, Ted and Balakrishna, Ashwin and Nair, Suraj and Rafailov, Rafael and Foster, Ethan P and Sanketi, Pannag R and Vuong, Quan and others},
    booktitle    = {Conference on Robot Learning},
    pages        = {2679--2713},
    year         = {2025},
    organization = {PMLR}
}

@inproceedings{black2025pi,
    title     = {\${\textbackslash}pi\_\{0.5\}\$: a Vision-Language-Action Model with Open-World Generalization},
    author    = {Kevin Black and Noah Brown and James Darpinian and Karan Dhabalia and Danny Driess and Adnan Esmail and Michael Robert Equi and Chelsea Finn and Niccolo Fusai and Manuel Y. Galliker and Dibya Ghosh and Lachy Groom and Karol Hausman and Brian Ichter and Szymon Jakubczak and Tim Jones and Liyiming Ke and Devin LeBlanc and Sergey Levine and Adrian Li-Bell and Mohith Mothukuri and Suraj Nair and Karl Pertsch and Allen Z. Ren and Lucy Xiaoyang Shi and Laura Smith and Jost Tobias Springenberg and Kyle Stachowicz and James Tanner and Quan Vuong and Homer Walke and Anna Walling and Haohuan Wang and Lili Yu and Ury Zhilinsky},
    booktitle = {9th Annual Conference on Robot Learning},
    year      = {2025}
}

@article{hassija2024interpreting,
    title     = {Interpreting black-box models: a review on explainable artificial intelligence},
    author    = {Hassija, Vikas and Chamola, Vinay and Mahapatra, Atmesh and Singal, Abhinandan and Goel, Divyansh and Huang, Kaizhu and Scardapane, Simone and Spinelli, Indro and Mahmud, Mufti and Hussain, Amir},
    journal   = {Cognitive Computation},
    volume    = {16},
    number    = {1},
    pages     = {45--74},
    year      = {2024},
    publisher = {Springer}
}

@article{kroemer2021review,
    title   = {A review of robot learning for manipulation: Challenges, representations, and algorithms},
    author  = {Kroemer, Oliver and Niekum, Scott and Konidaris, George},
    journal = {Journal of Machine Learning Research},
    volume  = {22},
    number  = {30},
    pages   = {1--82},
    year    = {2021}
}

@article{glanois2024survey,
    title     = {A survey on interpretable reinforcement learning},
    author    = {Glanois, Claire and Weng, Paul and Zimmer, Matthieu and Li, Dong and Yang, Tianpei and Hao, Jianye and Liu, Wulong},
    journal   = {Machine Learning},
    volume    = {113},
    number    = {8},
    pages     = {5847--5890},
    year      = {2024},
    publisher = {Springer}
}

@inproceedings{wen2025skilltree,
    title     = {SkillTree: Explainable Skill-Based Deep Reinforcement Learning for Long-Horizon Control Tasks},
    author    = {Wen, Yongyan and Li, Siyuan and Zuo, Rongchang and Yuan, Lei and Mao, Hangyu and Liu, Peng},
    booktitle = {Proceedings of the AAAI Conference on Artificial Intelligence},
    volume    = {39},
    number    = {20},
    pages     = {21491--21500},
    year      = {2025}
}

@inproceedings{paleja2022learning,
    title     = {Learning Interpretable, High-Performing Policies for Autonomous Driving},
    author    = {Paleja, Rohan and Niu, Yaru and Silva, Andrew and Ritchie, Chace and Choi, Sugju and Gombolay, Matthew},
    booktitle = {Robotics science and systems},
    year      = {2022}
}

@inproceedings{marton2025mitigating,
    title     = {Mitigating Information Loss in Tree-Based Reinforcement Learning via Direct Optimization},
    author    = {Marton, Sascha and Grams, Tim and Vogt, Florian and L{\"u}dtke, Stefan and Bartelt, Christian and Stuckenschmidt, Heiner},
    booktitle = {The Thirteenth International Conference on Learning Representations},
    year      = {2025}
}

@article{edmonds2019tale,
    title     = {A tale of two explanations: Enhancing human trust by explaining robot behavior},
    author    = {Edmonds, Mark and Gao, Feng and Liu, Hangxin and Xie, Xu and Qi, Siyuan and Rothrock, Brandon and Zhu, Yixin and Wu, Ying Nian and Lu, Hongjing and Zhu, Song-Chun},
    journal   = {Science Robotics},
    volume    = {4},
    number    = {37},
    pages     = {eaay4663},
    year      = {2019},
    publisher = {American Association for the Advancement of Science}
}

@inproceedings{greydanus2018visualizing,
    title        = {Visualizing and understanding atari agents},
    author       = {Greydanus, Samuel and Koul, Anurag and Dodge, Jonathan and Fern, Alan},
    booktitle    = {International Conference on Machine Learning},
    pages        = {1792--1801},
    year         = {2018},
    organization = {PMLR}
}

@inproceedings{koh2020concept,
    title        = {Concept bottleneck models},
    author       = {Koh, Pang Wei and Nguyen, Thao and Tang, Yew Siang and Mussmann, Stephen and Pierson, Emma and Kim, Been and Liang, Percy},
    booktitle    = {International Conference on Machine Learning},
    pages        = {5338--5348},
    year         = {2020},
    organization = {PMLR}
}

@inproceedings{yuksekgonul2023post,
    title     = {Post-hoc Concept Bottleneck Models},
    author    = {Yuksekgonul, Mert and Wang, Maggie and Zou, James},
    year      = {2023},
    booktitle = {The Eleventh International Conference on Learning Representations}
}

@inproceedings{oikarinen2023label,
    title     = {Label-free Concept Bottleneck Models},
    author    = {Oikarinen, Tuomas and Das, Subhro and Nguyen, Lam M and Weng, Tsui-Wei},
    year      = {2023},
    booktitle = {The Eleventh International Conference on Learning Representations}
}

@article{srivastava2024vlg,
    title   = {VLG-CBM: Training concept bottleneck models with vision-language guidance},
    author  = {Srivastava, Divyansh and Yan, Ge and Weng, Tsui-Wei},
    journal = {Advances in Neural Information Processing Systems},
    volume  = {37},
    pages   = {79057--79094},
    year    = {2024}
}

@inproceedings{hu2025semi,
    title     = {Semi-supervised concept bottleneck models},
    author    = {Hu, Lijie and Huang, Tianhao and Xie, Huanyi and Gong, Xilin and Ren, Chenyang and Hu, Zhengyu and Yu, Lu and Ma, Ping and Wang, Di},
    booktitle = {Proceedings of the IEEE/CVF International Conference on Computer Vision},
    pages     = {2110--2119},
    year      = {2025}
}

@inproceedings{liu2025hybrid,
    title     = {Hybrid concept bottleneck models},
    author    = {Liu, Yang and Zhang, Tianwei and Gu, Shi},
    booktitle = {Proceedings of the Computer Vision and Pattern Recognition Conference},
    pages     = {20179--20189},
    year      = {2025}
}

@inproceedings{wabartha2024piecewise,
    title     = {Piecewise linear parametrization of policies: Towards interpretable deep reinforcement learning},
    author    = {Wabartha, Maxime and Pineau, Joelle},
    booktitle = {The Twelfth International Conference on Learning Representations},
    year      = {2024}
}

@article{dhebar2022toward,
    title     = {Toward interpretable-AI policies using evolutionary nonlinear decision trees for discrete-action systems},
    author    = {Dhebar, Yashesh and Deb, Kalyanmoy and Nageshrao, Subramanya and Zhu, Ling and Filev, Dimitar},
    journal   = {IEEE Transactions on Cybernetics},
    volume    = {54},
    number    = {1},
    pages     = {50--62},
    year      = {2022},
    publisher = {IEEE}
}

@inproceedings{kohler2024interpretable,
    title     = {Interpretable and Editable Programmatic Tree Policies for Reinforcement Learning},
    author    = {Kohler, Hector and Delfosse, Quentin and Akrour, Riad and Kersting, Kristian and Preux, Philippe},
    booktitle = {Workshop on Interpretable Policies in Reinforcement Learning@ RLC-2024},
    year      = {2024}
}

@inproceedings{luo2024end,
    title        = {End-to-End Neuro-Symbolic Reinforcement Learning with Textual Explanations},
    author       = {Luo, Lirui and Zhang, Guoxi and Xu, Hongming and Yang, Yaodong and Fang, Cong and Li, Qing},
    booktitle    = {International Conference on Machine Learning},
    pages        = {33533--33557},
    year         = {2024},
    organization = {PMLR}
}

@article{bastani2018verifiable,
    title   = {Verifiable reinforcement learning via policy extraction},
    author  = {Bastani, Osbert and Pu, Yewen and Solar-Lezama, Armando},
    journal = {Advances in Neural Information Processing Systems},
    volume  = {31},
    year    = {2018}
}

@inproceedings{liu2024grounding,
    title        = {Grounding dino: Marrying dino with grounded pre-training for open-set object detection},
    author       = {Liu, Shilong and Zeng, Zhaoyang and Ren, Tianhe and Li, Feng and Zhang, Hao and Yang, Jie and Jiang, Qing and Li, Chunyuan and Yang, Jianwei and Su, Hang and others},
    booktitle    = {European Conference on Computer Vision},
    pages        = {38--55},
    year         = {2024},
    organization = {Springer}
}

@inproceedings{radford2021learning,
    title        = {Learning transferable visual models from natural language supervision},
    author       = {Radford, Alec and Kim, Jong Wook and Hallacy, Chris and Ramesh, Aditya and Goh, Gabriel and Agarwal, Sandhini and Sastry, Girish and Askell, Amanda and Mishkin, Pamela and Clark, Jack and others},
    booktitle    = {International conference on machine learning},
    pages        = {8748--8763},
    year         = {2021},
    organization = {PmLR}
}

@inproceedings{lee2024concept,
    title        = {Concept-based explanations in computer vision: Where are we and where could we go?},
    author       = {Lee, Jae Hee and Mikriukov, Georgii and Schwalbe, Gesina and Wermter, Stefan and Wolter, Diedrich},
    booktitle    = {European Conference on Computer Vision},
    pages        = {266--287},
    year         = {2024},
    organization = {Springer}
}

@inproceedings{wan2021nbdt,
    title     = {NBDT: Neural-Backed Decision Tree},
    author    = {Wan, Alvin and Dunlap, Lisa and Ho, Daniel and Yin, Jihan and Lee, Scott and Petryk, Suzanne and Bargal, Sarah Adel and Gonzalez, Joseph E},
    booktitle = {International Conference on Learning Representations},
    year      = {2021}
}

@article{dhebar2020interpretable,
    title     = {Interpretable rule discovery through bilevel optimization of split-rules of nonlinear decision trees for classification problems},
    author    = {Dhebar, Yashesh and Deb, Kalyanmoy},
    journal   = {IEEE Transactions on Cybernetics},
    volume    = {51},
    number    = {11},
    pages     = {5573--5584},
    year      = {2020},
    publisher = {IEEE}
}

@article{milani2024explainable,
    title     = {Explainable reinforcement learning: A survey and comparative review},
    author    = {Milani, Stephanie and Topin, Nicholay and Veloso, Manuela and Fang, Fei},
    journal   = {ACM Computing Surveys},
    volume    = {56},
    number    = {7},
    pages     = {1--36},
    year      = {2024},
    publisher = {ACM New York, NY}
}

@article{bai2025towards,
    title   = {Towards a unified understanding of robot manipulation: A comprehensive survey},
    author  = {Bai, Shuanghao and Song, Wenxuan and Chen, Jiayi and Ji, Yuheng and Zhong, Zhide and Yang, Jin and Zhao, Han and Zhou, Wanqi and Zhao, Wei and Li, Zhe and others},
    journal = {arXiv preprint arXiv:2510.10903},
    year    = {2025}
}

@inproceedings{ding2023task,
    title        = {Task and motion planning with large language models for object rearrangement},
    author       = {Ding, Yan and Zhang, Xiaohan and Paxton, Chris and Zhang, Shiqi},
    booktitle    = {2023 IEEE/RSJ International Conference on Intelligent Robots and Systems (IROS)},
    pages        = {2086--2092},
    year         = {2023},
    organization = {IEEE}
}

@inproceedings{du2024video,
    title     = {Video Language Planning},
    author    = {Du, Yilun and Yang, Sherry and Florence, Pete and Xia, Fei and Wahid, Ayzaan and Sermanet, Pierre and Yu, Tianhe and Abbeel, Pieter and Tenenbaum, Joshua B and Kaelbling, Leslie Pack and others},
    booktitle = {The Twelfth International Conference on Learning Representations},
    year      = {2024}
}

@inproceedings{song2023llm,
    title     = {LLM-Planner: Few-shot grounded planning for embodied agents with large language models},
    author    = {Song, Chan Hee and Wu, Jiaman and Washington, Clayton and Sadler, Brian M and Chao, Wei-Lun and Su, Yu},
    booktitle = {Proceedings of the IEEE/CVF international conference on computer vision},
    pages     = {2998--3009},
    year      = {2023}
}

@inproceedings{brohan2023can,
    title        = {Do as I can, not as I say: Grounding language in robotic affordances},
    author       = {Brohan, Anthony and Chebotar, Yevgen and Finn, Chelsea and Hausman, Karol and Herzog, Alexander and Ho, Daniel and Ibarz, Julian and Irpan, Alex and Jang, Eric and Julian, Ryan and others},
    booktitle    = {Conference on robot learning},
    pages        = {287--318},
    year         = {2023},
    organization = {PMLR}
}

@article{wu2023embodied,
    title   = {Embodied task planning with large language models},
    author  = {Wu, Zhenyu and Wang, Ziwei and Xu, Xiuwei and Lu, Jiwen and Yan, Haibin},
    journal = {arXiv preprint arXiv:2307.01848},
    year    = {2023}
}

@inproceedings{puri2020explain,
    title     = {Explain Your Move: Understanding Agent Actions Using Specific and Relevant Feature Attribution},
    author    = {Puri, Nikaash and Verma, Sukriti and Gupta, Piyush and Kayastha, Dhruv and Deshmukh, Shripad and Krishnamurthy, Balaji and Singh, Sameer},
    booktitle = {International Conference on Learning Representations},
    year      = {2020}
}

@inproceedings{tanno19adaptive,
    title     = {Adaptive Neural Trees},
    author    = {Tanno, Ryutaro and Arulkumaran, Kai and Alexander, Daniel and Criminisi, Antonio and Nori, Aditya},
    booktitle = {Proceedings of the 36th International Conference on Machine Learning},
    pages     = {6166--6175},
    year      = {2019},
    editor    = {Chaudhuri, Kamalika and Salakhutdinov, Ruslan},
    volume    = {97},
    series    = {Proceedings of Machine Learning Research},
    month     = {09--15 Jun},
    publisher = {PMLR}
}

@article{lu2024dynamic,
  title={A dynamic movement primitives-based tool use skill learning and transfer framework for robot manipulation},
  author={Lu, Zhenyu and Wang, Ning and Yang, Chenguang},
  journal={IEEE Transactions on Automation Science and Engineering},
  volume={22},
  pages={1748--1763},
  year={2024},
  publisher={IEEE}
}

@article{sobrin2025generating,
  title={Generating explanations for autonomous robots: a systematic review},
  author={Sobr{\'\i}n-Hidalgo, David and Guerrero-Higueras, {\'A}ngel Manuel and Matell{\'a}n-Olivera, Vicente},
  journal={IEEE Access},
  year={2025},
  publisher={IEEE}
}

@inproceedings{wong2021leveraging,
  title={Leveraging sparse linear layers for debuggable deep networks},
  author={Wong, Eric and Santurkar, Shibani and Madry, Aleksander},
  booktitle={International Conference on Machine Learning},
  pages={11205--11216},
  year={2021},
  organization={PMLR}
}

@inproceedings{he2016deep,
  title={Deep residual learning for image recognition},
  author={He, Kaiming and Zhang, Xiangyu and Ren, Shaoqing and Sun, Jian},
  booktitle={Proceedings of the IEEE conference on computer vision and pattern recognition},
  pages={770--778},
  year={2016}
}

@article{pedregosa2011scikit,
  title={Scikit-learn: Machine learning in Python},
  author={Pedregosa, Fabian and Varoquaux, Ga{\"e}l and Gramfort, Alexandre and Michel, Vincent and Thirion, Bertrand and Grisel, Olivier and Blondel, Mathieu and Prettenhofer, Peter and Weiss, Ron and Dubourg, Vincent and others},
  journal={the Journal of machine Learning research},
  volume={12},
  pages={2825--2830},
  year={2011},
  publisher={JMLR. org}
}

@inproceedings{deng2009imagenet,
  title={Imagenet: A large-scale hierarchical image database},
  author={Deng, Jia and Dong, Wei and Socher, Richard and Li, Li-Jia and Li, Kai and Fei-Fei, Li},
  booktitle={2009 IEEE conference on computer vision and pattern recognition},
  pages={248--255},
  year={2009},
  organization={Ieee}
}

%%%%%%%%%%%%%%%%%%%%%%%%%%%%%%%%%%%%%%%%%%%%%%%%%%%%%%%%%%%%

\clearpage
\appendix

\section{Concept Set Construction}\label{sec.concept_construction}
To construct a task-specific concept set, we query an LLM (e.g., GPT-5.4) with the task description, skill library, and expert execution sequence. The goal is not to obtain generic visual categories, but to identify binary predicates that are observable from the robot's visual input and relevant to high-level skill selection. We focus on three types of concepts:
\begin{itemize}
  \item \textbf{Object presence:} whether a task-relevant object is visible in the scene.
  \item \textbf{Object state:} whether an object has a relevant state, such as open, closed, filled, or empty.
  \item \textbf{Spatial relation:} whether a task-relevant relation holds between objects, such as inside, on, or near.
\end{itemize}
The generated concepts are then used as the predicate set for VLM-based concept annotation. We denote the finalized concept set as $\mathcal{C}=\{c^{(1)},\dots,c^{(M)}\}$, where $c^{(k)}$ is the $k$-th semantic concept. The prompt is shown below.

\begin{plainpromptbox}
Prompt for concept set construction.

You are constructing a concept set for a robot manipulation task.

Your goal is to generate a fixed set of visual concepts based on given SKILL LIBRARY.

Important notes:
- Concepts must follow a strict naming convention
- Only use the allowed concept types listed below
- Do NOT invent new concept types
- Concepts must be visually observable from a single image

Skill library: [SKILL LIBRARY of the task]

Concept types:
1. Visibility
Format: <object>_visible

2. Spatial Relation (object-object or object-location)
Format: <object>_on_surface, <object>_on_counter, <object>_on_plate, <object>_inside_<container>, <object>_above_<object>
Examples: cup_on_surface, funnel_above_pot, milk_bottle_inside_fridge

3. Container State (binary content state)
Format: <object>_empty, <object>_contains_liquid, <object>_contains_dark_liquid
Examples: cup_empty, coffee_cup_contains_liquid, pot_contains_liquid

4. Object State / Configuration
Format: <object>_upright, <object>_tilted
Examples: gooseneck_kettle_upright, gooseneck_kettle_tilted

5. Interaction Proxy (only if visually obvious)
Format: <object>_pouring_to_<object>
Example: gooseneck_kettle_pouring_to_funnel

6. Appliance State
Format: <appliance>_door_open
Examples: fridge_door_open, cabinet_door_open, microwave_door_open

7. Generic Object Group (category abstraction)
Format: any_object_on_counter, any_object_inside_fridge, any_object_inside_cabinet, object_inside_<container>, object_on_<location>

8. Shape / Appearance Category (only predefined categories)
Format: round_<color>_object_on_counter, elongated_<color>_object_on_counter, small_<color>_box_on_counter
Examples: round_orange_object_on_counter, elongated_red_object_on_counter, small_light_box_on_counter

Output Requirements:
- Return ALL concepts following the formats above
- Do NOT include explanations
- Do NOT generate concepts outside these categories
- Avoid duplicates
\end{plainpromptbox}

\subsection{Manual Validation and Final Concept Sets}
The LLM-generated concepts are treated as candidate concepts rather than the final concept set. We manually review the candidates using predefined criteria: concepts must be visually observable from the side-camera image, relevant to high-level skill selection, expressible as binary predicates, and non-redundant with other concepts. During this step, we remove ambiguous, non-visual, task-irrelevant, or duplicate concepts. The finalized concept sets are fixed before model training and \textbf{shared across ConceptTree and all concept-based baselines} to ensure a fair comparison.

To reduce annotation cost, we exploit the temporal structure of the collected skill demonstrations. For each skill execution, the raw dataset contains 50 side-camera observations captured after the skill starts (Appendix \ref{app.data_collection_eval}). We request VLM annotations only for the first observation in each 50-frame window and assign the resulting concept labels to the remaining observations in the same window. This keeps the number of VLM queries small while providing concept supervision for all training frames. Although some later observations may occasionally deviate from the first frame in concept semantics, we found this approximation to work well in practice for training the concept layer. Table \ref{tbl.final_concepts} lists the finalized concept sets used in our experiments.

\begin{table}[htbp]
  \centering
  \small
  \caption{\textbf{Finalized task-specific concept sets.}}
  \label{tbl.final_concepts}
  \newcommand{\conceptname}[1]{\texttt{\detokenize{#1}}}
  \newcommand{\block}[1]{%
    \begin{minipage}[t]{\linewidth}
      \raggedright\footnotesize #1
    \end{minipage}}
  \begin{tabular}{p{0.15\linewidth}p{0.78\linewidth}}
    \toprule
    Task & Concept Set \\
    \midrule
    \textit{Fruits-Snacks}
    & \block{
    \conceptname{round_orange_object_on_counter};
    \conceptname{round_yellow_object_on_counter};
    \conceptname{round_green_object_on_counter};
    \conceptname{elongated_red_object_on_counter};
    \conceptname{elongated_yellow_object_on_counter};
    \conceptname{small_red_box_on_counter};
    \conceptname{small_light_box_on_counter};
    \conceptname{round_object_inside_fridge};
    \conceptname{elongated_object_inside_fridge};
    \conceptname{round_object_inside_cabinet};
    \conceptname{elongated_object_inside_cabinet};
    \conceptname{box_object_inside_cabinet};
    \conceptname{any_object_on_counter};
    \conceptname{any_object_inside_fridge};
    \conceptname{any_object_inside_cabinet};
    \conceptname{fridge_door_open};
    \conceptname{cabinet_door_open}
    } \\
    \addlinespace
    \textit{Heat-Bread}
    & \block{
    \conceptname{microwave_door_open};
    \conceptname{bread_visible};
    \conceptname{bread_on_surface};
    \conceptname{bread_inside_microwave};
    \conceptname{bread_on_plate};
    \conceptname{plate_visible};
    \conceptname{plate_on_surface};
    \conceptname{plate_contains_object};
    \conceptname{object_inside_microwave};
    \conceptname{object_on_plate};
    \conceptname{object_on_counter}
    } \\
    \addlinespace
    \textit{Cola}
    & \block{
    \conceptname{fridge_open};
    \conceptname{cup_visible};
    \conceptname{cup_on_surface};
    \conceptname{cup_inside_fridge};
    \conceptname{cup_contains_liquid};
    \conceptname{can_visible};
    \conceptname{can_on_surface};
    \conceptname{bin_visible};
    \conceptname{object_inside_bin};
    \conceptname{milk_bottle_visible};
    \conceptname{milk_bottle_on_surface};
    \conceptname{milk_bottle_inside_fridge};
    \conceptname{object_inside_fridge};
    \conceptname{object_on_surface}
    } \\
    \addlinespace
    \textit{Coffee}
    & \block{
    \conceptname{funnel_visible};
    \conceptname{pot_visible};
    \conceptname{kettle_visible};
    \conceptname{gooseneck_kettle_visible};
    \conceptname{cup_visible};
    \conceptname{coffee_cup_visible};
    \conceptname{coffee_powder_visible};
    \conceptname{funnel_above_pot};
    \conceptname{funnel_above_table_surface};
    \conceptname{cup_empty};
    \conceptname{cup_contains_liquid};
    \conceptname{coffee_cup_empty};
    \conceptname{cup_contains_dark_liquid};
    \conceptname{coffee_cup_contains_liquid};
    \conceptname{coffee_cup_contains_dark_liquid};
    \conceptname{funnel_empty};
    \conceptname{pot_contains_liquid};
    \conceptname{pot_empty};
    \conceptname{gooseneck_kettle_upright};
    \conceptname{gooseneck_kettle_tilted};
    \conceptname{gooseneck_kettle_pouring_to_funnel}
    } \\
    \bottomrule
  \end{tabular}
\end{table}

\subsection{VLM-based Concept Annotation}
Given the task-specific concept set, we construct binary concept labels for each visual observation using GPT-5.1 as an offline verifier. For each image, the VLM evaluates the full list of task-relevant concepts and returns a JSON object with one binary value per concept, where 1 indicates that the predicate holds in the scene and 0 otherwise. We denote the binary value assigned to concept $c^{(k)}$ for observation $o_t$ as $y_t(c^{(k)})$. The complete VLM-derived concept target vector is $\mathbf{y}_t =
\big(y_t(c^{(1)}),\dots,y_t(c^{(M)})\big)^\top \in \{0,1\}^M$. This vector is used as supervision for training the concept layer.

\begin{plainpromptbox}
You are a visual reasoning assistant.

You are given:
1) An input image.
2) A list of concepts.

Your task:
For EACH concept, determine whether it is TRUE (1) or FALSE (0) based ONLY on what is clearly visible in the image.

IMPORTANT RULES:
- Only use visual evidence from the image.
- Do NOT guess or assume anything not clearly visible.
- If uncertain, output 0.
- Be strict and conservative.
- Each concept must be evaluated independently.

Concept definitions:
- "on_surface":
  an object is clearly placed on a visible flat supporting surface.

- "inside_container":
  an object is clearly enclosed within a container-like structure, where the boundary of the container is visible.

- "container_open":
  a container has a visible opening, such that its interior is partially visible.

- "object_visible":
  at least one object is clearly visible in the scene.

- "object_near_boundary":
  an object is close to the edge of a surface or container opening.

Output format (STRICT):
Return ONLY a valid JSON object with NO explanation.

Format:
{{
  "concept_1": 0 or 1,
  "concept_2": 0 or 1,
  ...
}}

Now evaluate the following concepts:

{concept_list}
\end{plainpromptbox}

\section{Training Details}
All experiments are run on a workstation with an Intel Core i9-14900KF CPU with 128GB RAM and a single NVIDIA RTX 4090 GPU. In our setting, the training time is very short because the dataset and model are relatively small, with the concept layer and the decision tree training taking less than 2 minutes for each task.

\paragraph{Concept layer training.}
For visual feature extraction, we use a ResNet-18 \cite{he2016deep} backbone pretrained on ImageNet-1K \cite{deng2009imagenet} as a fixed encoder. Each preprocessed RGB observation is mapped to a 512-dimensional image feature. The training details are summarized in Table \ref{tbl.concept_training}.

\begin{table}[htbp]
  \centering
  \caption{\textbf{Training configuration for the ConceptTree concept layer.}}
  \label{tbl.concept_training}
  \begin{tabular}{lc}
    \toprule
    Hyperparameter & Value \\
    \midrule
    Visual encoder & ResNet-18 \\
    Image feature dimension & 512 \\
    Input resolution & $256\times256$ \\
    Optimizer & Adam \\
    Learning rate & 1e-3 \\
    Batch size & 512 \\
    Training steps & 1000 \\
    \bottomrule
  \end{tabular}
\end{table}

\paragraph{Decision tree training.}
We use a single axis-aligned decision tree implemented with \texttt{sklearn.tree.DecisionTreeClassifier} \cite{pedregosa2011scikit}. The decision tree is trained on the learned concept $\mathbf{c}_t$. For the no-history setting, the tree is trained directly on the concept values of the training observations and their corresponding skill labels. For the history setting, we construct an expanded training set by pairing history frames with current frames. Specifically, for each skill decision, we take the Cartesian product between the frames associated with the previous skill and the frames associated with the current skill, and concatenate their concept values as the tree input. Since the first skill has no previous skill context, we instead form the Cartesian product between its own frames and itself. At inference time, we use the same convention. For the first decision step in the history setting, the current concept vector is copied as the history input. Each paired sample is labeled with the current skill. The tree depth is controlled by a maximum depth parameter $d$, the split quality is measured by information gain using the entropy criterion, and class imbalance is handled with balanced class weights. Deeper trees are more expressive, but they reduce interpretability. We select a fixed depth of 7 for all tasks based on ablation analysis, which provides a good balance between performance and interpretability. In practice, we find that $d=4$ works well for most tasks.

\section{Ablation Study on Skill Loss Weight}
We ablate the skill-aware loss weight $\lambda\in \{0, 0.05, 0.1, 0.2, 0.5, 1.0\}$ with the history input and evaluate each weight across tree depths $d\in [4,10]$. For each task and weight, we report the best-performing depth in Table \ref{tbl.skillw_ablation}. Overall, introducing a nonzero skill-aware loss consistently improves over the w/o skill loss baseline. The task-level behavior is non-uniform. On \textit{Coffee}, which exhibits the strongest visual ambiguity, larger skill-aware weights are most beneficial, the completion rate improves from $44.0\%$ at $\lambda=0$ to $70.91\%$ at $\lambda=0.2$ and further to $80.36\%$ at $\lambda=1.0$. In contrast, \textit{Cola} prefers much smaller weights, the best result is already obtained at $\lambda=0.05$ ($95.33\%$), while larger weights substantially degrade performance (e.g., $73.78\%$ at $\lambda=0.5$). \textit{Heat-Bread} is best at $\lambda=0.1$, whereas \textit{Fruits-Snacks} peaks at $\lambda=0.2$ with perfect test performance. These results suggest that the skill-aware loss primarily acts as a representation-shaping mechanism. Stronger skill supervision is especially useful in tasks where stage boundaries are visually ambiguous, as in \textit{Coffee}, but it can over-distort the concept space in easier tasks where the representation is already sufficiently separable. We set $\lambda=0.1$ for \textit{Fruits-Snacks}, \textit{Heat-Bread}, and \textit{Cola} and $\lambda=0.5$ for \textit{Coffee} in the main results.

\begin{table}[htbp]
  \centering
  \caption{\textbf{Skill-aware loss weight ablation under the history input setting.} We report completion rate with mean$\pm$std in \%, and the best depth $d$ for the corresponding task and weight.}
  \label{tbl.skillw_ablation}
  \setlength{\tabcolsep}{2pt}
  \small
  \begin{tabular}{lcccc}
    \toprule
    Parameter &
    \textit{Fruits-Snacks} &
    \textit{Heat-Bread} &
    \textit{Cola} &
    \textit{Coffee} \\
    \midrule
    $\lambda=0.0$  & $95.56\pm5.54$ ($d=4$) & $78.00\pm2.84$ ($d=4$) & $90.00\pm0.00$ ($d=7$) & $44.00\pm12.35$ ($d=10$)\\
    $\lambda=0.05$ & $97.14\pm2.54$ ($d=4$) & $76.86\pm1.40$ ($d=5$) & $\mathbf{95.33\pm3.17}$ ($d=7$) & $58.91\pm10.45$ ($d=10$)\\
    $\lambda=0.1$ & $94.92\pm6.22$ ($d=4$) & $\mathbf{84.57\pm4.64}$ ($d=5$) & $95.11\pm4.48$ ($d=4$) & $58.55\pm6.34$ ($d=10$)\\
    $\lambda=0.2$ & $\mathbf{100.00\pm0.00}$ ($d=4$) & $80.57\pm10.29$ ($d=5$) & $90.44\pm4.61$ ($d=6$) & $70.91\pm11.33$ ($d=10$) \\
    $\lambda=0.5$ & $96.83\pm4.38$ ($d=4$) & $78.29\pm17.94$ ($d=4$) & $73.78\pm5.10$ ($d=8$) & $66.55\pm14.25$ ($d=10$) \\
    $\lambda=1.0$ & $92.38\pm4.86$ ($d=4$)  & $74.00\pm14.04$ ($d=4$) & $75.11\pm4.13$ ($d=8$) & $\mathbf{80.36\pm5.06}$ ($d=10$) \\
    \bottomrule
  \end{tabular}
\end{table}

\section{Concept Importance Analysis}
We analyze the interpretability of the decision tree by tracing the concepts that appear along test-time decision paths. For each test step, we record the concepts used by the executed path, normalize their frequencies within that path, and then average the normalized contributions over the entire test set. Importantly, we perform the analysis at the level of individual episode steps (skills) rather than globally merging all occurrences of the same skill, since the same skill can be invoked multiple times under different visual contexts. We compare the resulting concept importance distributions for ConceptTree w/ history and ConceptTree w/o history, using a fixed depth per task.

\begin{figure}[htbp]
    \centering
    \includegraphics[width=\textwidth]{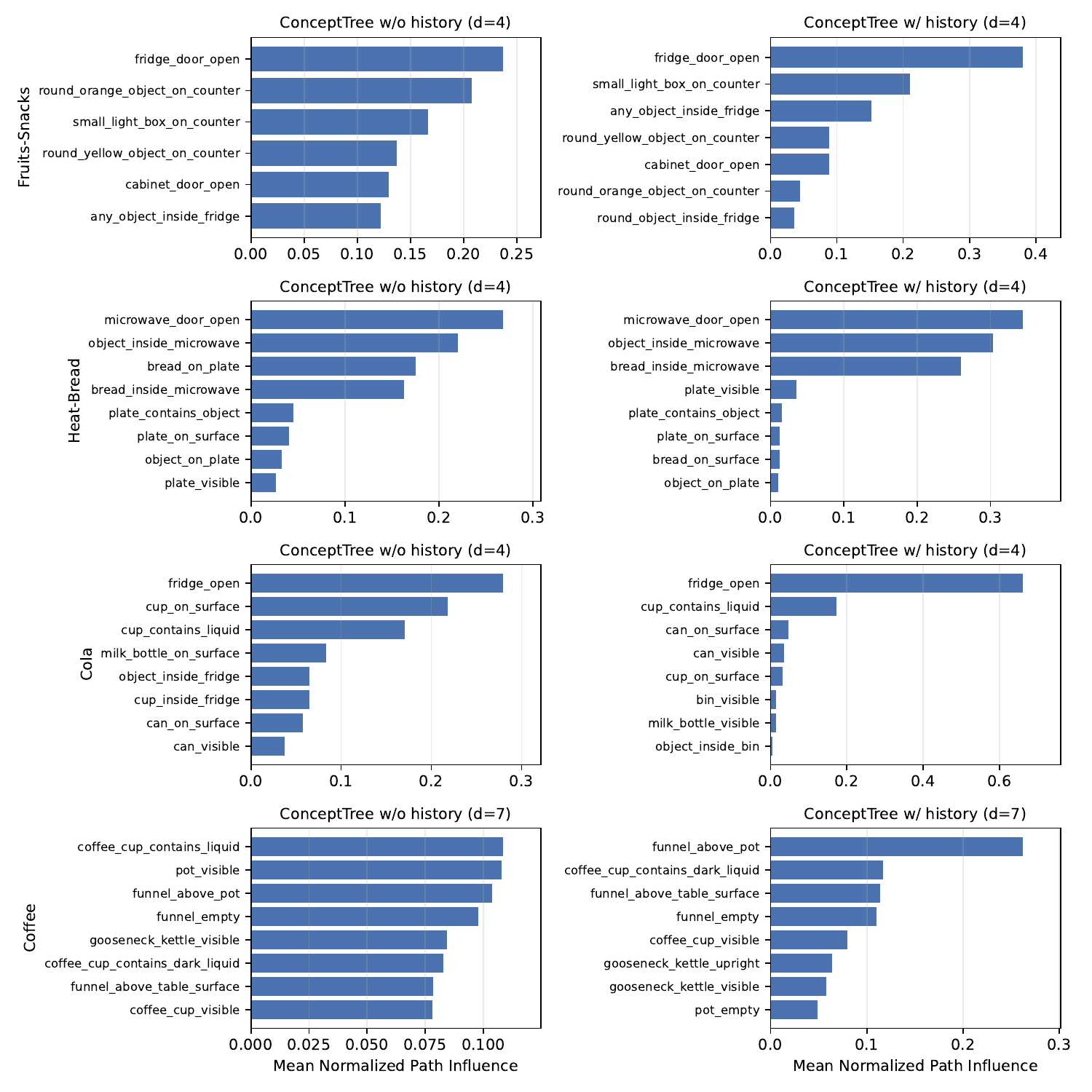}
    \caption{\textbf{Visualization of concept importance in decision paths across tasks.}}
    \label{fig.tree_importance}
\end{figure}

The results in Fig. \ref{fig.tree_importance} show that the most influential concepts are highly consistent across the history and no-history variants, and they align well with human intuition about which visual cues should matter for each task. In other words, the tree tends to rely on semantically meaningful concepts, such as object state, container state, or spatial relations, rather than obscure or obviously spurious signals. This consistency suggests that adding history improves temporal context without changing the semantic basis of the decision process. Therefore, the decision tree is interpretable not only because it exposes an explicit rule path, but also because the concepts that dominate its predictions remain stable and human-understandable across settings.

\section{Decision Tree Visualization}
\begin{figure}[htbp]
    \centering
    \includegraphics[width=\textwidth]{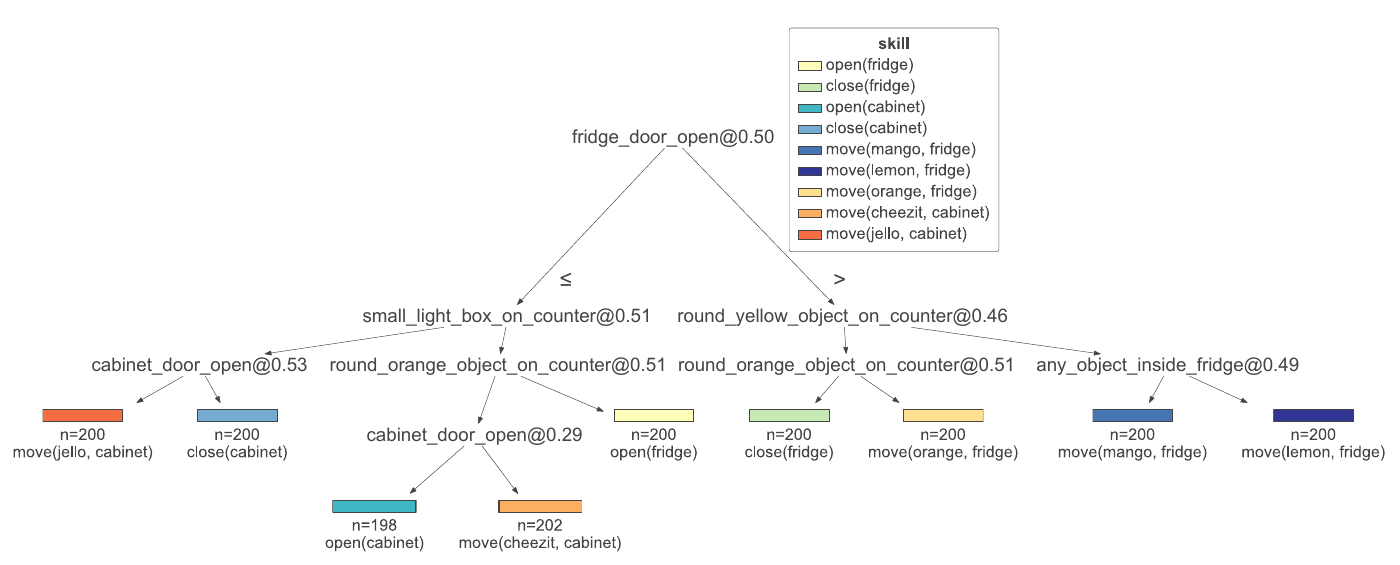}
    \caption{\textbf{Visualization of decision tree for the \textit{Fruits-Snacks} task.}}
    \label{fig.tree_fruits_snacks}
\end{figure}

\begin{figure}[htbp]
    \centering
    \includegraphics[width=\textwidth]{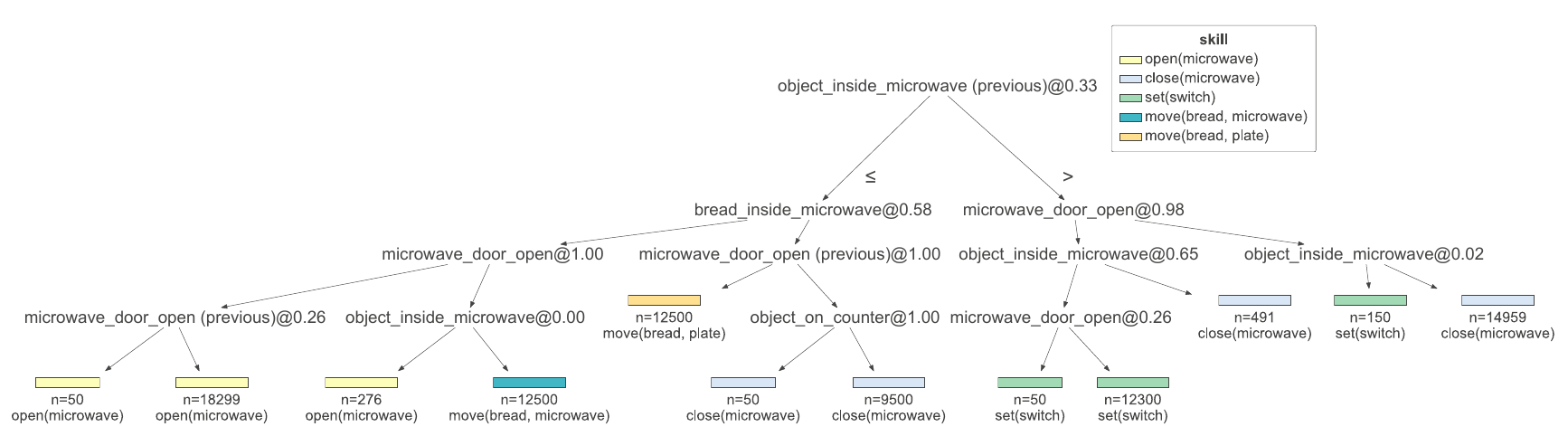}
    \caption{\textbf{Visualization of decision tree for the \textit{Heat-Bread} task.}}
    \label{fig.tree_heat_bread}
\end{figure}

\begin{figure}[htbp]
    \centering
    \includegraphics[width=\textwidth]{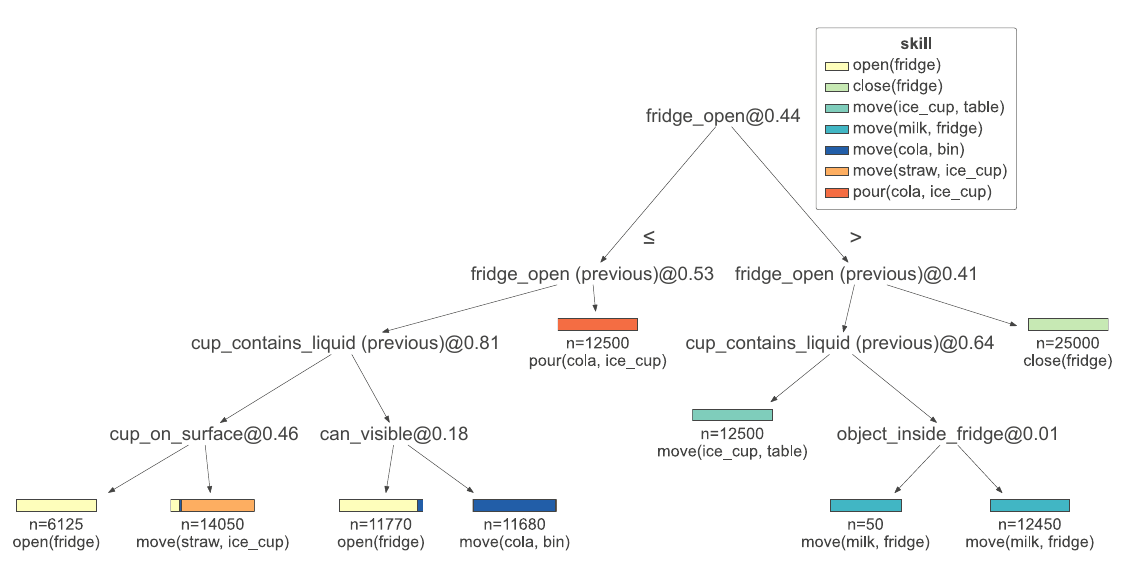}
    \caption{\textbf{Visualization of decision tree for the \textit{Cola} task.}}
    \label{fig.tree_cola}
\end{figure}

\begin{figure}[htbp]
    \centering
    \includegraphics[width=\textwidth]{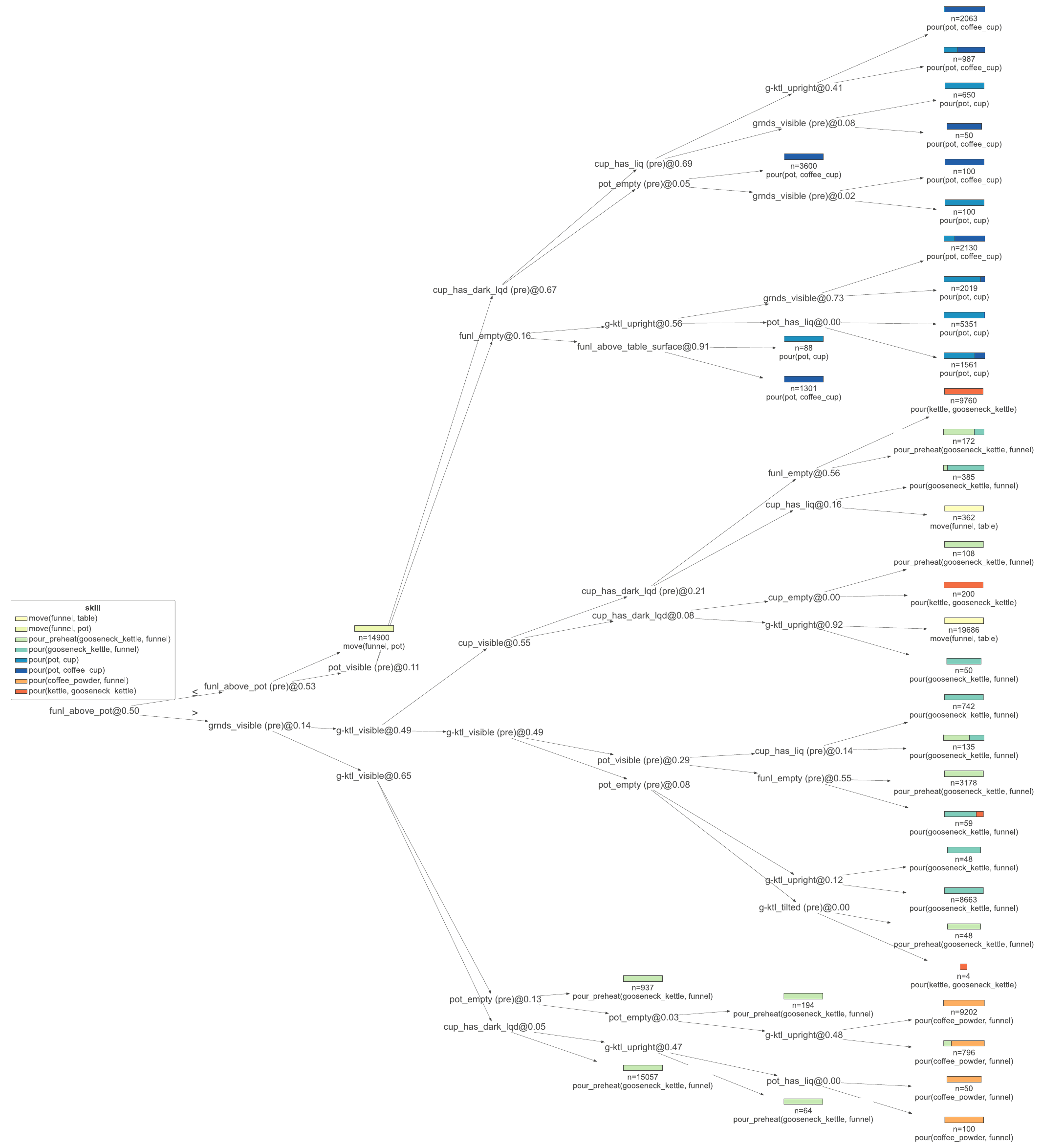}
    \caption{\textbf{Visualization of decision tree for the \textit{Coffee} task.}}
    \label{fig.tree_coffee}
\end{figure}

We visualized the learned decision trees for each task in Fig. \ref{fig.tree_fruits_snacks}, \ref{fig.tree_heat_bread}, \ref{fig.tree_cola} and \ref{fig.tree_coffee}. These visualizations expose the actual sequence of concept predicates used by the high-level policy and make it possible to inspect how different branches correspond to different skill choices.

\section{Decision Tree vs. Sparse Linear Model}\label{sec.dt_vs_linear}
We further compare the interpretability of a single decision tree and a sparse linear classifier \cite{wong2021leveraging} when both operate on the \textbf{same task-specific concept representation} for a fair comparison. For the tree model, we use the corresponding ConceptTree model and report complexity statistics at maximum depth $d=7$. For ConceptTree, we report local decision complexity, including UC (Used Concepts on test-time decisions), IN (Internal Nodes, standing for the number of parameters), and APL (Average Path Length). For the sparse linear model, we report global sparsity, including UC (Used Concepts), NZW (total NonZero Weights, indicating the effective parameter count), and CPS (Concepts Per Skill, i.e., the average number of nonzero concept coefficients for each skill class).

\begin{table}[htbp]
  \centering
  \caption{\textbf{Interpretability statistics for ConceptTree and the sparse linear classifier.} ConceptTree statistics measure concepts actually used along test-time decision paths, while sparse linear statistics measure global nonzero weights in the classifier.}
  \label{tbl.tree_linear_interpretability}
  \small
  \begin{tabular}{lcccccc}
    \toprule
    \multirow{2}{*}{Task}
    & \multicolumn{3}{c}{ConceptTree}
    & \multicolumn{3}{c}{Sparse Linear Model} \\
    \cmidrule(lr){2-4}\cmidrule(lr){5-7}
    & UC$\downarrow$ & IN$\downarrow$ & APL$\downarrow$ & UC$\downarrow$ & NZW$\downarrow$ & CPS$\downarrow$ \\
    \midrule
    \textit{Fruits-Snacks} & $5.40{\pm}0.49$ & $8.00{\pm}0.00$ & $3.23{\pm}0.02$ & $16.0{\pm}0.0$ & $143.0{\pm}2.2$ & $15.7{\pm}0.2$ \\
    \textit{Heat-Bread} & $5.60{\pm}1.02$ & $12.60{\pm}1.36$ & $3.81{\pm}0.19$ & $11.0{\pm}0.0$ & $95.6{\pm}1.9$ & $19.0{\pm}0.4$ \\
    \textit{Cola} & $8.40{\pm}1.36$ & $17.60{\pm}0.49$ & $4.09{\pm}0.09$ & $14.0{\pm}0.0$ & $119.2{\pm}1.5$ & $17.3{\pm}0.2$ \\
    \textit{Coffee} & $14.00{\pm}1.10$ & $55.40{\pm}4.50$ & $5.84{\pm}0.09$ & $13.4{\pm}0.8$ & $152.2{\pm}7.2$ & $18.7{\pm}0.9$ \\
    \bottomrule
  \end{tabular}
\end{table}

The tree model uses a small number of concepts at test time, and each prediction corresponds to a short rule chain. As shown in Table \ref{tbl.tree_linear_interpretability}, ConceptTree uses fewer concepts than the sparse linear model on \textit{Fruits-Snacks}, \textit{Heat-Bread}, and \textit{Cola}, while keeping the average decision path within roughly three to four predicates. On the more difficult \textit{Coffee} task, the number of active concepts becomes comparable, but the tree still produces a local path of only $5.84{\pm}0.09$ predicates on average. In contrast, the sparse linear classifier relies on distributed evidence, with 95-152 nonzero weights and roughly 15--17 concept coefficients per skill across tasks. This indicates that global sparsity in the parameter matrix does not necessarily translate into a concise explanation for an individual prediction.

We further perform a concept-budget analysis for the sparse linear classifier by retaining only the top-$K$ globally important concepts and masking the rest. The results show that sparse linear performance usually requires a relatively large concept set. Completion rate is recovered only with about 12 concepts on \textit{Cola}, nearly all 11 concepts on \textit{Heat-Bread}, and 15 concepts on \textit{Coffee}, while the easier \textit{Fruits-Snacks} task is recovered with the top 7 concepts. This suggests that sparse linear models remain globally sparse but rely on broadly distributed concept evidence, especially as task complexity increases. For comparison, ConceptTree uses a much smaller set of concepts while achieving better performance.

\begin{figure}[htbp]
    \centering
    \includegraphics[width=\textwidth]{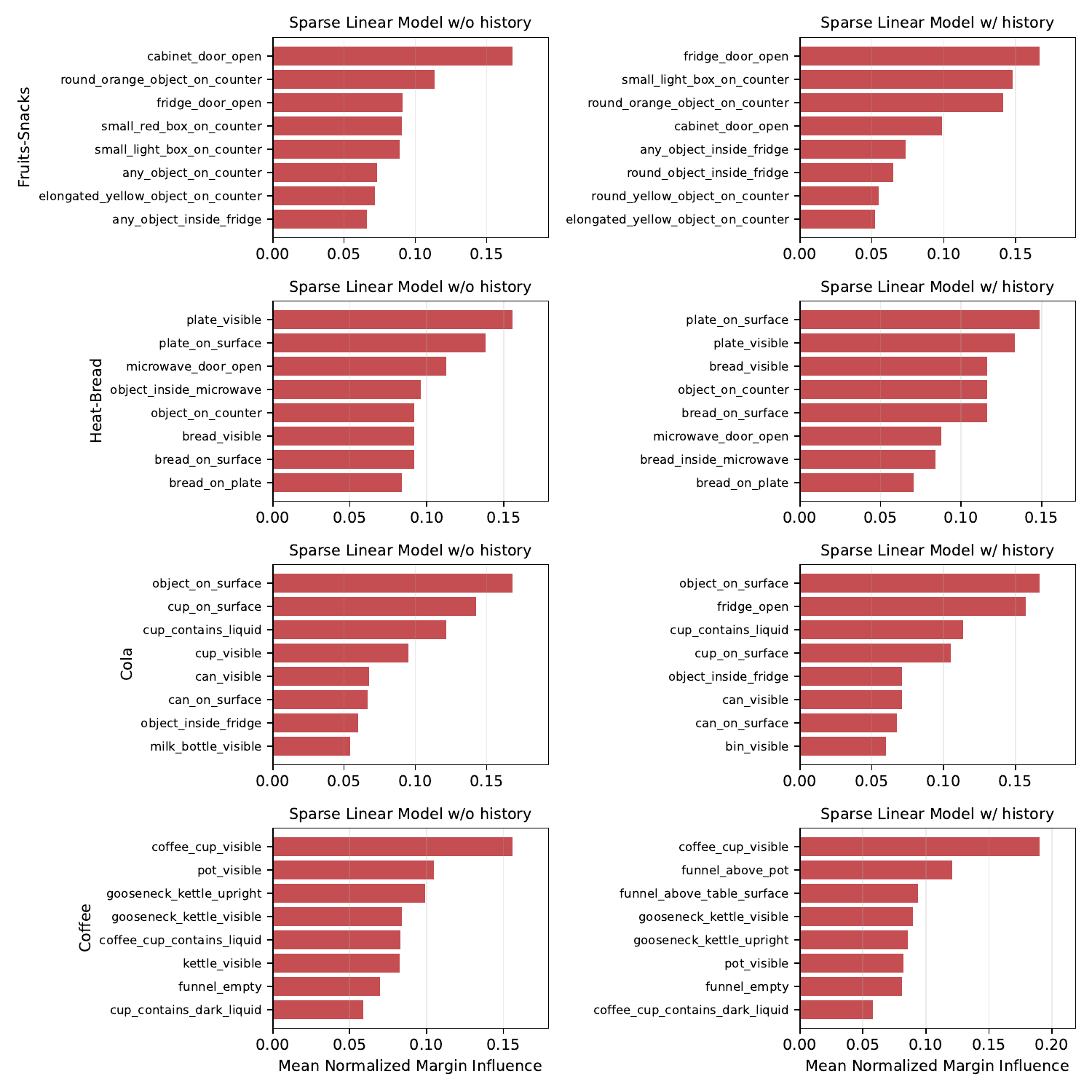}
    \caption{\textbf{Visualization of top-8 concept weights in sparse linear models across tasks.}}
    \label{fig.linear_importance}
\end{figure}

\paragraph{Concept influence comparison.}
We analyze test-time concept influence for both model families. For ConceptTree, the influential concepts remain highly consistent between the history and no-history variants, as shown in Fig. \ref{fig.tree_importance}, and align with intuitive task cues such as object state and spatial relations. The tree also exhibits sharper influence distributions, where a few dominant concepts explain most of the decision path. In contrast, the sparse linear weights in Fig. \ref{fig.linear_importance} are spread more evenly across multiple concepts, suggesting that its predictions are supported by distributed evidence rather than a compact set of decisive predicates. These results suggest that the tree bases its decisions on stable and semantically meaningful concepts while providing a more localized decision path than the linear model.

Overall, the two models provide different forms of interpretability. A sparse linear classifier can be sparse at the parameter level, but its predictions are supported by distributed weighted evidence and do not yield a concise local reasoning chain. This is limiting for robotic sequential decision making, where each skill choice affects later observations and failures must be traced to the specific scene condition that caused an incorrect branch in the task. In contrast, ConceptTree exposes a short concept-level path for each prediction, making the selected skill attributable to a small and stable chain of semantic predicates. Therefore, compared with a sparse linear model, the tree is better suited to interpretable high-level robotic decision making: it not only uses concepts selectively, but also provides stronger local interpretability for step-wise diagnosis and intervention.

\section{Additional Intervention Case Study}\label{sec.additional_intervention}
This section describes how we identify the concept to intervene on when ConceptTree predicts an incorrect skill. Rather than inspecting all concepts, we use the tree structure to find the first predicate where the erroneous decision path diverges from a nearby path that predicts the correct skill.

\paragraph{Path-based diagnosis.}
For an incorrect prediction $a_t=g(\tilde{\mathbf{c}}_t)$, let $\rho_{\text{err}}=(n_0,\ldots,n_m)$ be the traversed root-to-leaf path, where $n_0$ is the root node, $n_m$ is the reached leaf node, and each internal node $n_j$ tests a predicate $d_{n_j}(\tilde{\mathbf{c}}_t)$. Since the same skill may appear in multiple leaves, the ground-truth skill $a_t^*$ can correspond to multiple correct paths:
\begin{equation}
  \mathcal{R}(a_t^*) = \{\rho: \mathrm{leaf}(\rho) \text{ predicts } a_t^*\}.
\end{equation}
We select the correct path that shares the longest prefix with the erroneous path:
\begin{equation}
  \rho_{\text{near}} = \arg\max_{\rho\in\mathcal{R}(a_t^*)}\mathrm{LCP}(\rho_{\text{err}}, \rho),
\end{equation}
where $\mathrm{LCP}$ denotes the number of shared nodes from the root. Equivalently, this finds the correct-skill leaf with the deepest lowest common ancestor with the erroneous leaf. The last shared internal node is the nearest divergent parent node, whose split predicate identifies the concept dimension and threshold that send the two paths to different branches.

\paragraph{Intervention target.}
Suppose the nearest divergent parent node tests dimension $k$ of $\tilde{\mathbf{c}}_t$ with threshold $\tau$. If $\rho_{\text{err}}$ and $\rho_{\text{near}}$ follow different branches at this node, then dimension $k$ is the first semantic variable that prevents the current prediction from reaching the nearest correct path. We use the corresponding concept as the intervention target. In the no-history setting, this dimension corresponds to a concept value in $\mathbf{c}_t$; in the history setting, it further indicates whether the concept belongs to $\mathbf{c}_{t-1}$ or $\mathbf{c}_t$. We then edit only this concept value, move it to the semantically correct side of the split, and recompute the tree prediction without changing model parameters. This provides a local counterfactual diagnosis: it identifies the first concept-level predicate that must change to redirect the decision toward the correct skill.

\begin{figure}
    \centering
    \includegraphics[width=\textwidth]{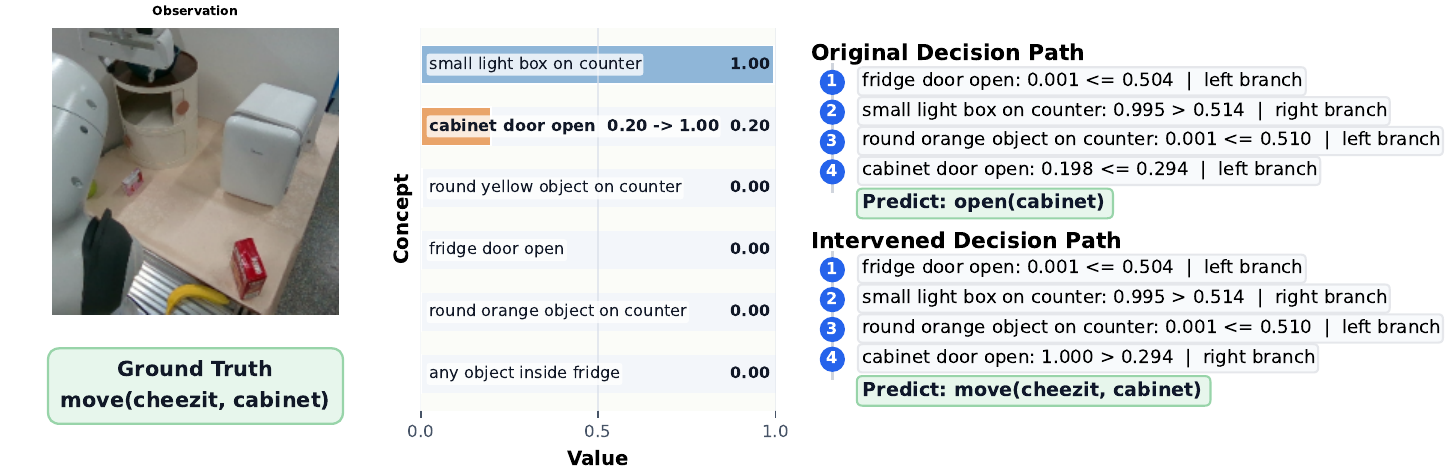}
    \caption{\textbf{Visualization of the intervention case study on \textit{Fruits-Snacks}.}}
    \label{fig.intervention_fruits}
\end{figure}

\begin{figure}
    \centering
    \includegraphics[width=\textwidth]{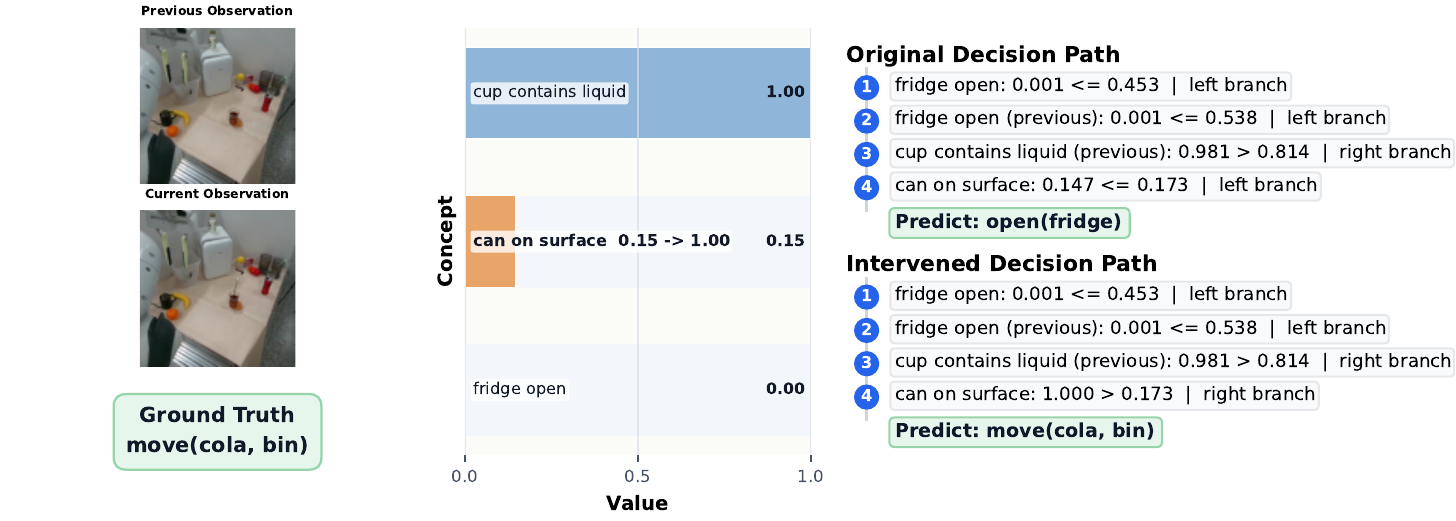}
    \caption{\textbf{Visualization of the intervention case study on \textit{Cola}.}}
    \label{fig.intervention_cola}
\end{figure}

\begin{figure}
    \centering
    \includegraphics[width=\textwidth]{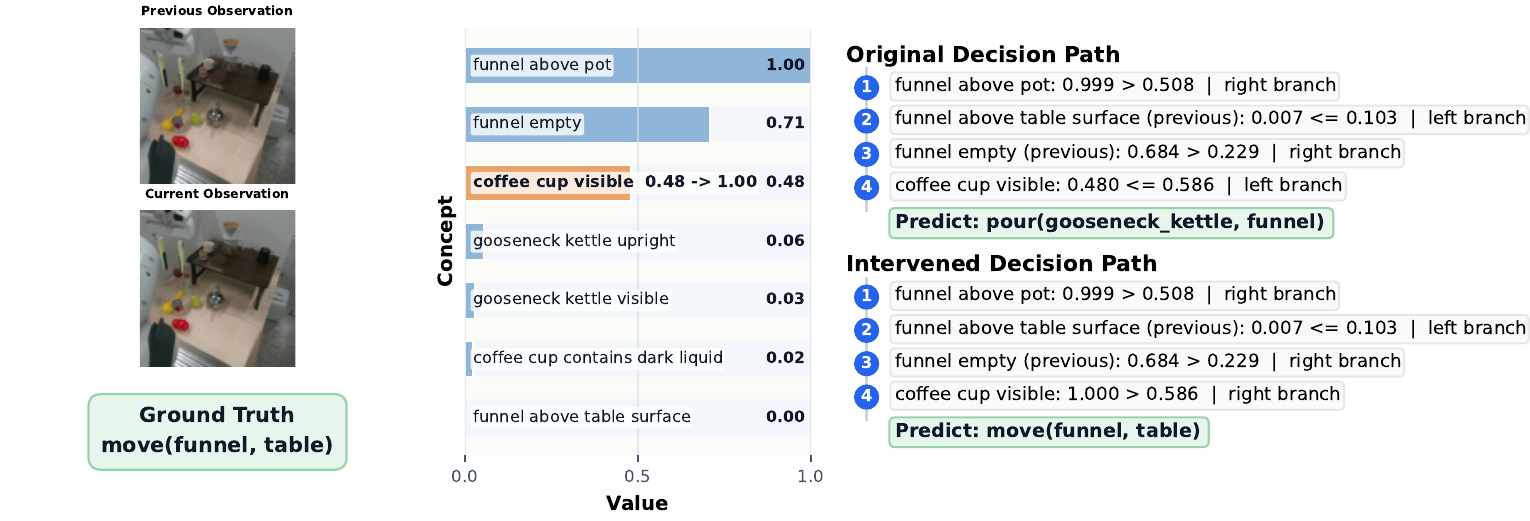}
    \caption{\textbf{Visualization of the intervention case study on \textit{Coffee}.}}
    \label{fig.intervention_coffee}
\end{figure}

\paragraph{Additional examples.}
Figs. \ref{fig.intervention_fruits}, \ref{fig.intervention_cola}, and \ref{fig.intervention_coffee} show additional intervention examples. In \textit{Fruits-Snacks}, the model incorrectly predicts \texttt{open(cabinet)} instead of \texttt{move(cheezit, cabinet)} because the concept value of \texttt{cabinet\_door\_open} is 0.20, which does not match the actual observation and the threshold (0.29). Changing this single concept redirects the path to the correct skill. The \textit{Cola} and \textit{Coffee} examples follow the same procedure, illustrating how tree paths narrow failure analysis to a specific concept predicate rather than requiring a search over the full concept set.

\section{Implementation Details}
In this section, we provide details on the robotic task setup used in our experiments. The task suite, low-level skill library, and physical evaluation episodes are fixed across all compared methods, while the high-level decision module evaluated in this paper is ConceptTree rather than a black-box recurrent or transformer policy.

\subsection{Robot Platform and Low-level Skills}
All experiments are conducted on a physical Franka Emika Panda 7-DOF robot arm with a stationary side camera that provides a global view of the workspace. The high-level policy selects from a predefined library of discrete skills, and each selected skill is executed by a pretrained low-level controller. These low-level controllers correspond to reusable motion primitives such as \texttt{open}, \texttt{close}, \texttt{move}, and \texttt{pour}. They are trained with the Human-in-the-Loop Sample Efficient Robotic Reinforcement Learning (HIL-SERL) framework \cite{luo2025precise}, which incorporates human demonstrations and real-time corrective interventions during policy learning. In our experiments, the low-level skills are kept fixed across all methods so that evaluation focuses on high-level skill selection.

\subsection{Tasks and Skill Sequences}
\begin{figure}[htpb]
    \centering
    \includegraphics[width=\textwidth]{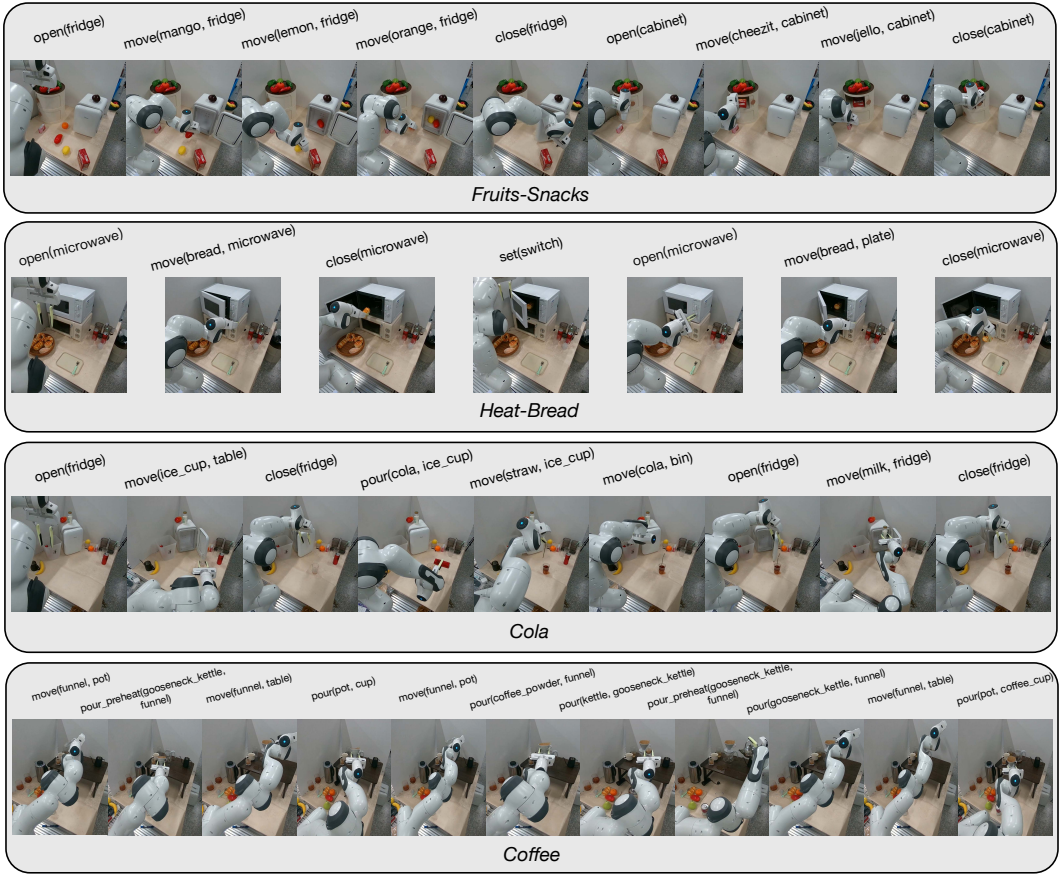}
    \caption{\textbf{Illustrations of the tasks in the experiments and skill sequences.}}
    \label{fig.visualization}
\end{figure}

We evaluate on four long-horizon manipulation tasks: \textit{Fruits-Snacks}, \textit{Heat-Bread}, \textit{Cola}, and \textit{Coffee}. Each task is decomposed into a fixed task-relevant skill library and a corresponding expert execution sequence, as shown in Table \ref{tbl.tasks}. The tasks are designed to cover different sources of difficulty, including temporal dependency, repeated skill usage, fine-grained object interactions, and sensitivity to visual details. Fig. \ref{fig.visualization} illustrates the task scenes and skill sequences.

\subsection{Low-level Action Policy Details}
The low-level action policies provide the executable primitives used by all high-level decision methods. Each skill policy is trained with HIL-SERL using human demonstrations and corrective interventions, and maps proprioceptive and visual observations to continuous end-effector commands. In our platform, robust vision-based skills can be obtained with a success rate exceeding 90\% within approximately 1 hour of real-world training. To maintain consistency between consecutive skills, we use scripted resets to return the robot arm to a standard observation pose after each skill execution, preventing the arm or gripper from occluding the side-camera view used for high-level decision making.

\subsection{Preprocessing}
RGB observations are captured from the side camera at 10 Hz with an original resolution of $1280\times 720$ pixels. For high-level visual decision making, each raw side-camera image is center-cropped to remove peripheral workspace regions and resized to $256\times256$ pixels. Concept annotations are constructed over these preprocessed observations, and all compared high-level decision methods operate on the same observation stream and skill labels.

\subsection{Data Collection and Evaluation}\label{app.data_collection_eval}
During training data collection, the robot follows the expert skill sequence for each task, and each observation is associated with the current high-level skill label. Rather than using the entire trajectory of a low-level skill execution, we keep the first 50 side-camera frames after each skill starts. This window captures the visual conditions under which the high-level skill choice is made, while avoiding later frames dominated by low-level execution transients, intermediate object motions, and post-action states that can deviate from the original decision context. In practice, this selective sampling increases visual diversity across skill executions while reducing label ambiguity for high-level decision learning. We collect 5 episodes for each task, resulting in datasets containing 2250, 1750, 2250, and 2750 frames for \textit{Fruits-Snacks}, \textit{Heat-Bread}, \textit{Cola}, and \textit{Coffee}, respectively.

For evaluation, we deploy each trained high-level policy on the physical robot under controlled variations in task-irrelevant object categories, object locations, and lighting conditions. At each decision point, the policy predicts the next skill from the current observation; if the prediction is incorrect, we record the error but execute the ground-truth skill before continuing the episode. This protocol evaluates high-level decisions under real robot observations while allowing later decisions in the same long-horizon episode to be measured. For each method, we run 10, 10, 10, and 5 evaluation episodes on \textit{Fruits-Snacks}, \textit{Heat-Bread}, \textit{Cola}, and \textit{Coffee}, respectively.

\begin{table}[htpb]
	\centering
	\caption{\textbf{Task settings used in the experiments.} For each task, we list the predefined skill library and the expert execution sequence used for high-level decision evaluation.}
	\label{tbl.tasks}
  \footnotesize
  \setlength{\tabcolsep}{5pt}
    \begin{tabular}{lll}
        \toprule
        Task & Skill Library & Execution Sequence \\
        \cmidrule{1-3}
        \multirow{9}{*}{\textit{Fruits-Snacks}} & \texttt{open(fridge)} & \texttt{open(fridge)} \\
        & \texttt{close(fridge)} & \texttt{move(mango, fridge)} \\
        & \texttt{open(cabinet)} & \texttt{move(lemon, fridge)} \\
        & \texttt{close(cabinet)} & \texttt{move(orange, fridge)} \\
        & \texttt{move(mango, fridge)} & \texttt{close(fridge)} \\
        & \texttt{move(lemon, fridge)} & \texttt{open(cabinet)} \\
        & \texttt{move(orange, fridge)} & \texttt{move(cheezit, cabinet)} \\
        & \texttt{move(cheezit, cabinet)} & \texttt{move(jello, cabinet)} \\
        & \texttt{move(jello, cabinet)} & \texttt{close(cabinet)} \\
        \cmidrule{1-3}
        \multirow{7}{*}{\textit{Heat-Bread}} & \multirow{7}{*}{\makecell[l]{\texttt{open(microwave)}\\\texttt{close(microwave)}\\\texttt{set(switch)}\\\texttt{move(bread, microwave)}\\\texttt{move(bread, plate)}}} & \texttt{open(microwave)} \\
        & & \texttt{move(bread, microwave)} \\
        & & \texttt{close(microwave)} \\
        & & \texttt{set(switch)} \\
        & & \texttt{open(microwave)} \\
        & & \texttt{move(bread, plate)} \\
        & & \texttt{close(microwave)} \\
        \cmidrule{1-3}
        \multirow{9}{*}{\textit{Cola}} & \multirow{9}{*}{\makecell[l]{\texttt{open(fridge)}\\\texttt{close(fridge)}\\\texttt{move(ice\_cup, table)}\\\texttt{move(milk, fridge)}\\\texttt{move(cola, bin)}\\\texttt{move(straw, ice\_cup)}\\\texttt{pour(cola, ice\_cup)}}} & \texttt{open(fridge)} \\
        & & \texttt{move(ice\_cup, table)} \\
        & & \texttt{close(fridge)} \\
        & & \texttt{pour(cola, ice\_cup)} \\
        & & \texttt{move(straw, ice\_cup)} \\
        & & \texttt{move(cola, bin)} \\
        & & \texttt{open(fridge)} \\
        & & \texttt{move(milk, fridge)} \\
        & & \texttt{close(fridge)} \\
        \cmidrule{1-3}
        \multirow{11}{*}{\textit{Coffee}} & \multirow{11}{*}{\makecell[l]{\texttt{move(funnel, table)}\\\texttt{move(funnel, pot)}\\\texttt{pour\_preheat(gooseneck\_kettle, funnel)}\\\texttt{pour(gooseneck\_kettle, funnel)}\\\texttt{pour(pot, cup)}\\\texttt{pour(pot, coffee\_cup)}\\\texttt{pour(coffee\_powder, funnel)}\\\texttt{pour(kettle, gooseneck\_kettle)}}} & \texttt{move(funnel, pot)} \\
        & & \texttt{pour\_preheat(gooseneck\_kettle, funnel)} \\
        & & \texttt{move(funnel, table)} \\
        & & \texttt{pour(pot, cup)} \\
        & & \texttt{move(funnel, pot)} \\
        & & \texttt{pour(coffee\_powder, funnel)} \\
        & & \texttt{pour(kettle, gooseneck\_kettle)} \\
        & & \texttt{pour\_preheat(gooseneck\_kettle, funnel)} \\
        & & \texttt{pour(gooseneck\_kettle, funnel)} \\
        & & \texttt{move(funnel, table)} \\
        & & \texttt{pour(pot, coffee\_cup)} \\
        \bottomrule
  \end{tabular}
\end{table}

\section{Additional Visualization of Episodes and Decision Paths}
\begin{figure}[htbp]
    \centering
    \includegraphics[width=\textwidth]{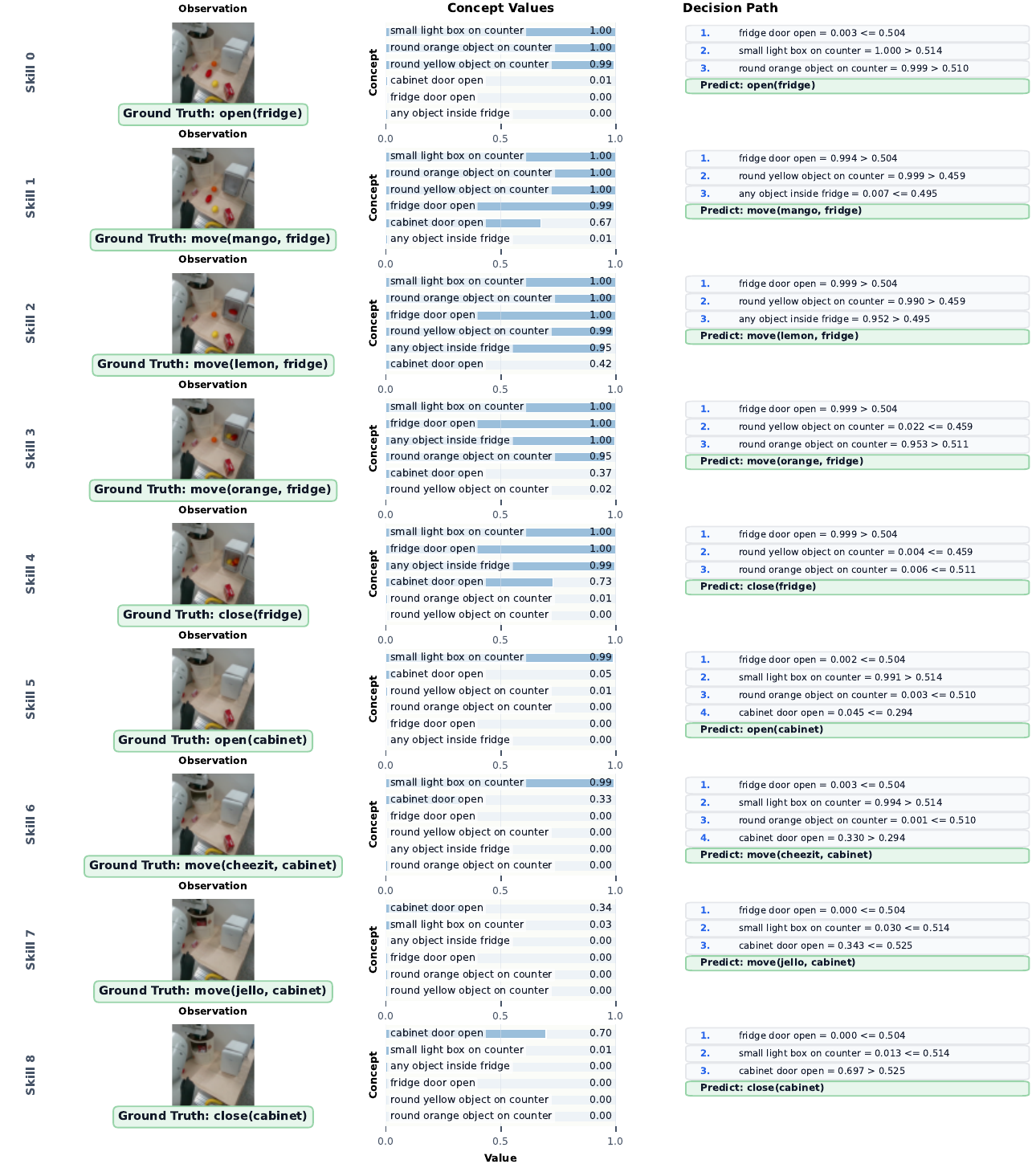}
    \caption{\textbf{Visualization of a test episode for the \textit{Fruits-Snacks}.}}
    \label{fig.episode_fruits_snacks}
\end{figure}

\begin{figure}[htbp]
    \centering
    \includegraphics[width=\textwidth]{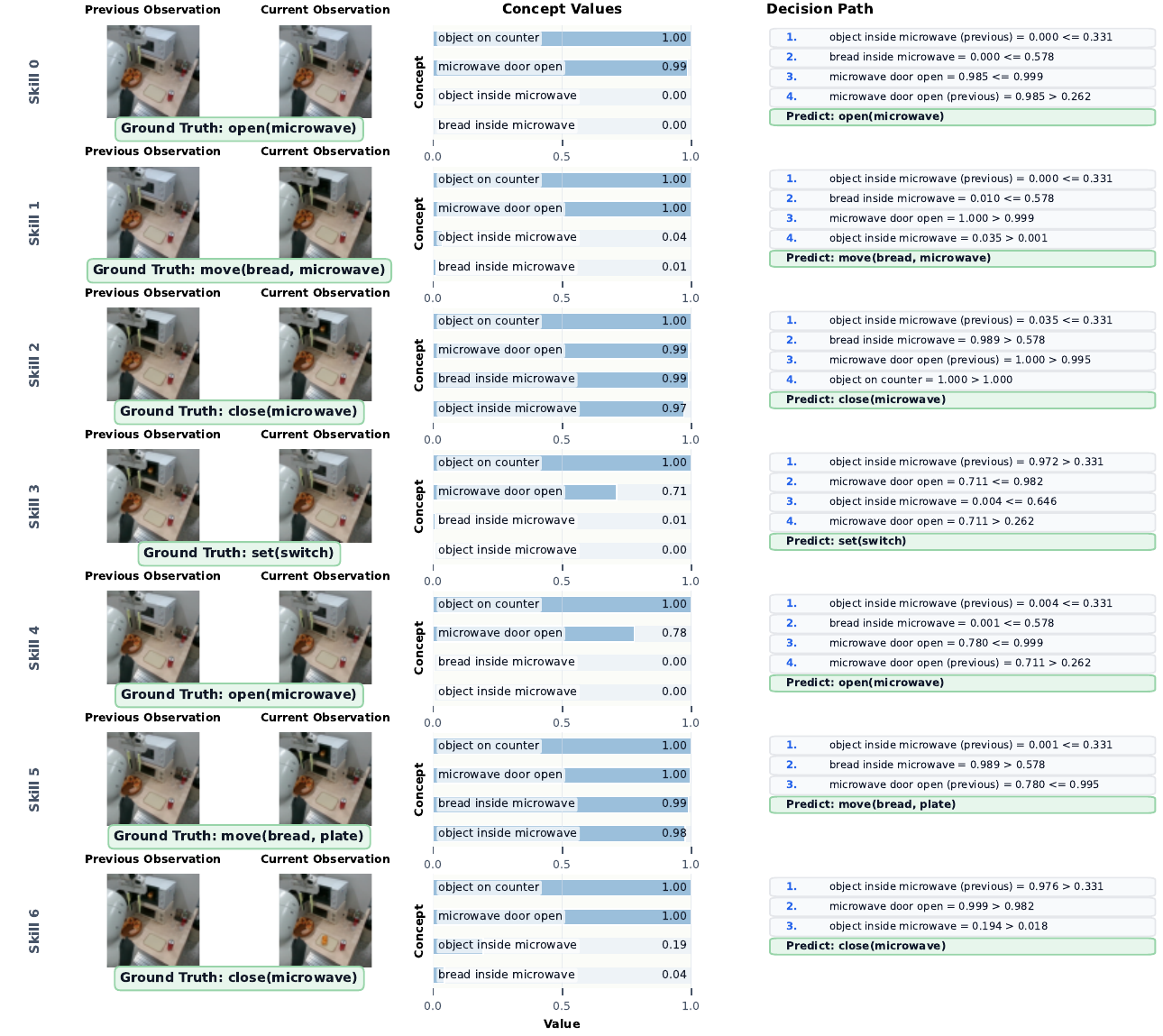}
    \caption{\textbf{Visualization of a test episode for the \textit{Heat-Bread} with history input.}}
    \label{fig.episode_heat_bread}
\end{figure}

\begin{figure}[htbp]
    \centering
    \includegraphics[width=\textwidth]{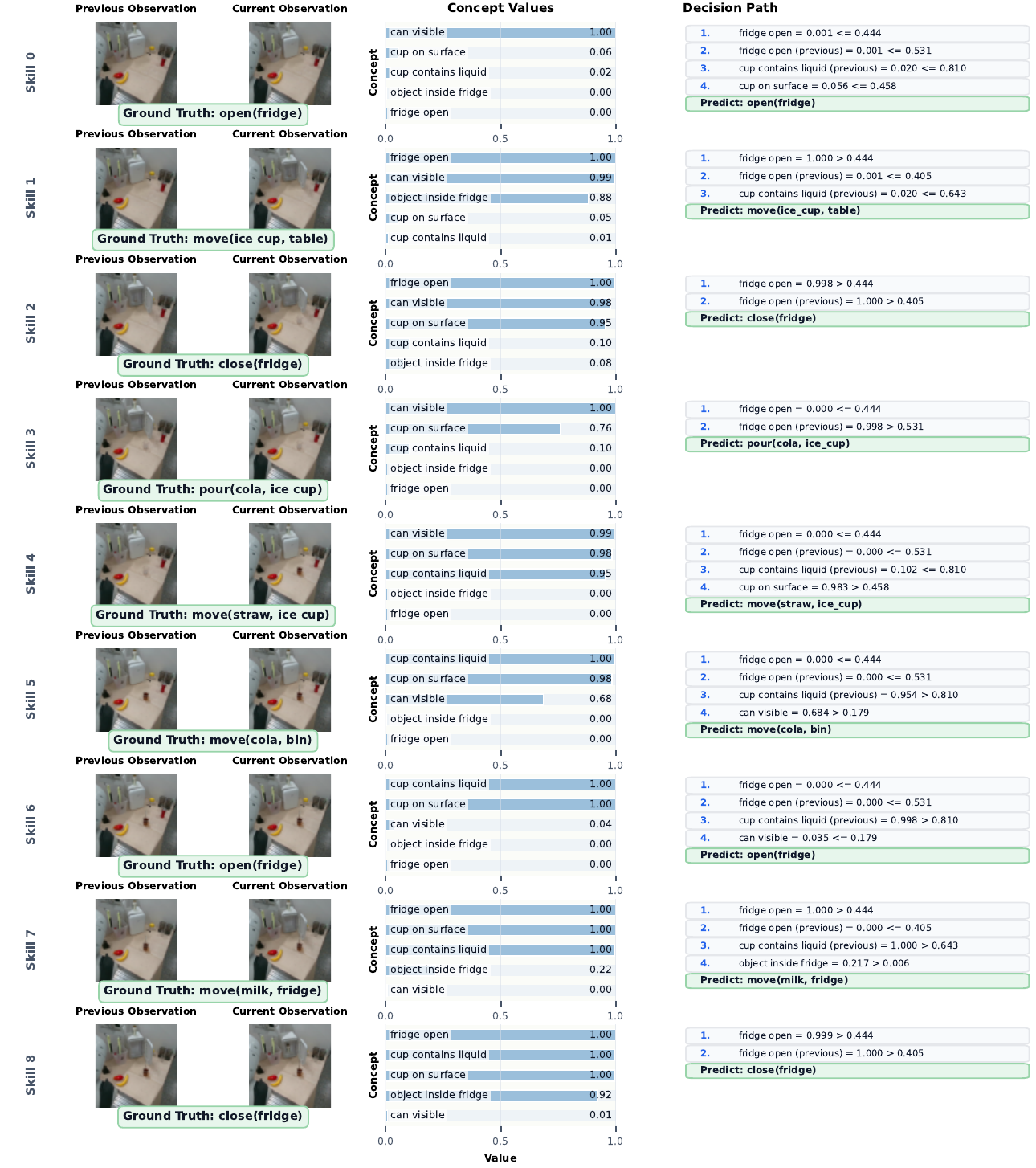}
    \caption{\textbf{Visualization of a test episode for the \textit{Cola} with history input.}}
    \label{fig.episode_cola}
\end{figure}

%%%%%%%%%%%%%%%%%%%%%%%%%%%%%%%%%%%%%%%%%%%%%%%%%%%%%%%%%%%%

% \clearpage
% \input{checklist.tex}

\end{document}